\documentclass[11pt]{article}

\PassOptionsToPackage{table}{xcolor}

\usepackage[preprint]{acl}

\usepackage{times}
\usepackage{latexsym}
\usepackage[T1]{fontenc}
\usepackage[utf8]{inputenc}
\usepackage{microtype}
\usepackage{inconsolata}
\usepackage{graphicx}

\usepackage{booktabs}
\usepackage{amsmath}
\usepackage{amssymb}
\usepackage{amsfonts}
\usepackage{multirow}
\usepackage{makecell}
\usepackage{array}
\usepackage{enumitem}
\usepackage{titletoc}
\titlecontents{lsection}[1.6em]{\addvspace{2pt}\bfseries\filright\hyphenpenalty=10000\exhyphenpenalty=10000}{\contentslabel{1.6em}}{\hspace*{-1.6em}}{\titlerule*[0.6pc]{.}\contentspage}
\titlecontents{lsubsection}[3.9em]{\filright\hyphenpenalty=10000\exhyphenpenalty=10000}{\contentslabel{2.3em}}{\hspace*{-2.3em}}{\titlerule*[0.6pc]{.}\contentspage}
\usepackage{algorithm}
\usepackage{algorithmic}

\definecolor{table_color}{HTML}{9C8AD0}
\definecolor{gaincolor}{HTML}{2E7D4F}
\newcommand{\gain}[1]{\textsubscript{\textcolor{gaincolor}{+#1}}}
\newtheorem{property}{Property}

\graphicspath{{figures/}}
\ifdefined\pdfsuppressptexinfo\pdfsuppressptexinfo=-1\relax\fi

\title{AdaGround: Token-Normalized Adaptive Zooming with Verifier-in-the-Loop Refinement for GUI Grounding}

\hypersetup{pdftitle={AdaGround: Token-Normalized Adaptive Zooming with Verifier-in-the-Loop Refinement for GUI Grounding},
  pdfauthor={Bingchao Wang, Yongcheng Jing, Chengyu Wang, Jun Huang, Jie Yang, Wei Liu, Dacheng Tao}}
\author{
  \textbf{Bingchao Wang}$^{1,2*}$,
  \textbf{Yongcheng Jing}$^{2}$,
  \textbf{Chengyu Wang}$^{3}$,
  \textbf{Jun Huang}$^{3}$, \\
  \textbf{Jie Yang}$^{1\dagger}$,
  \textbf{Wei Liu}$^{1\dagger}$,
  \textbf{Dacheng Tao}$^{2\dagger}$ \\[3pt]
  $^{1}$Shanghai Jiao Tong University \quad
  $^{2}$Nanyang Technological University \quad
  $^{3}$Alibaba Cloud Computing \\[2pt]
  \small\texttt{\{bc\_wang, jieyang, weiliucv\}@sjtu.edu.cn}, \texttt{yongcheng.jing@ntu.edu.sg}, \texttt{dacheng.tao@gmail.com}, \\
  \small\texttt{\{chengyu.wcy, huangjun.hj\}@alibaba-inc.com}}

\begin{document}
\maketitle
{\renewcommand{\thefootnote}{\fnsymbol{footnote}}%
\footnotetext[1]{This work is completed during Bingchao Wang's research attachment at NTU. $^{\dagger}$Corresponding Authors.}}
\begin{abstract}
GUI grounding on high-resolution screens fails mostly on small targets, and zooming in is a common training-free remedy.
However, most existing zooming pipelines do not match the zoom to the target's size, accept the zoomed answer unconditionally and never revisit a wrong answer.
To address these issues, we propose AdaGround, a training-free framework with three components:
(1)~\textbf{Token-Normalized Adaptive Zooming}, which sets the magnification, field of view and zoom depth of each sample from the model's own estimate of the target size;
(2)~\textbf{Verifier-in-the-Loop Refinement}, which uses one marked-click Yes/No verifier to keep or reject each zoom, choose between zoom chains, trigger re-grounding and, with an existence query, decide when to abstain;
and (3)~\textbf{Compositional Structure Grounding}, which composes spreadsheet cells from their headers.
On Qwen3-VL models, AdaGround surpasses state-of-the-art training-free zooming methods and many trained GUI models, improving accuracy over the backbone by 7.4 to 20.6 points on OSWorld-G and ScreenSpot-Pro, and its gains carry over to held-out UI-Vision.
Code is available at \url{https://github.com/bcwang-sjtu/AdaGround}.
\end{abstract}

\section{Introduction}
\label{sec:intro}

Autonomous agents that operate graphical user interfaces (GUIs) are central to generalist assistants~\citep{guiagentssurvey, mllmbasedguiagents, brainedgui}. Their perceptual core is \emph{GUI grounding}, which takes an instruction and a screenshot and localizes the target element precisely enough to act on it~\citep{gpt4twebagent, winspot}. Large vision-language models (VLMs)~\citep{qwen3vl2025, seedvl2025, claudecomputeruse2024} cast grounding as coordinate generation~\citep{seeclick2024, cogagent2024, uitars2025}, which remains fragile on professional, high-resolution software, where targets are small and dense~\citep{screenspotpro2025}. Training-based specialists~\citep{uground2025, osatlas2024, infigui2025, guig1_2025} and learned focusing~\citep{guispotlight2025, focusui2026} improve it but need annotated data and re-training per backbone. This makes \emph{training-free test-time refinement} attractive. It crops the screenshot around a coarse prediction and re-grounds at higher resolution.

\begin{figure}[t]
\centering
\includegraphics[width=\columnwidth]{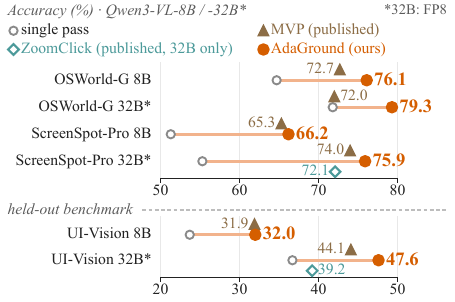}
\caption{Accuracy of AdaGround, the single pass and the published results of MVP and ZoomClick, each measured from its own single pass. ZoomClick reports Qwen3-VL-32B only. UI-Vision is held out.}
\label{fig:teaser}
\end{figure}
\begin{figure*}[t]
    \centering
    \includegraphics[width=\textwidth]{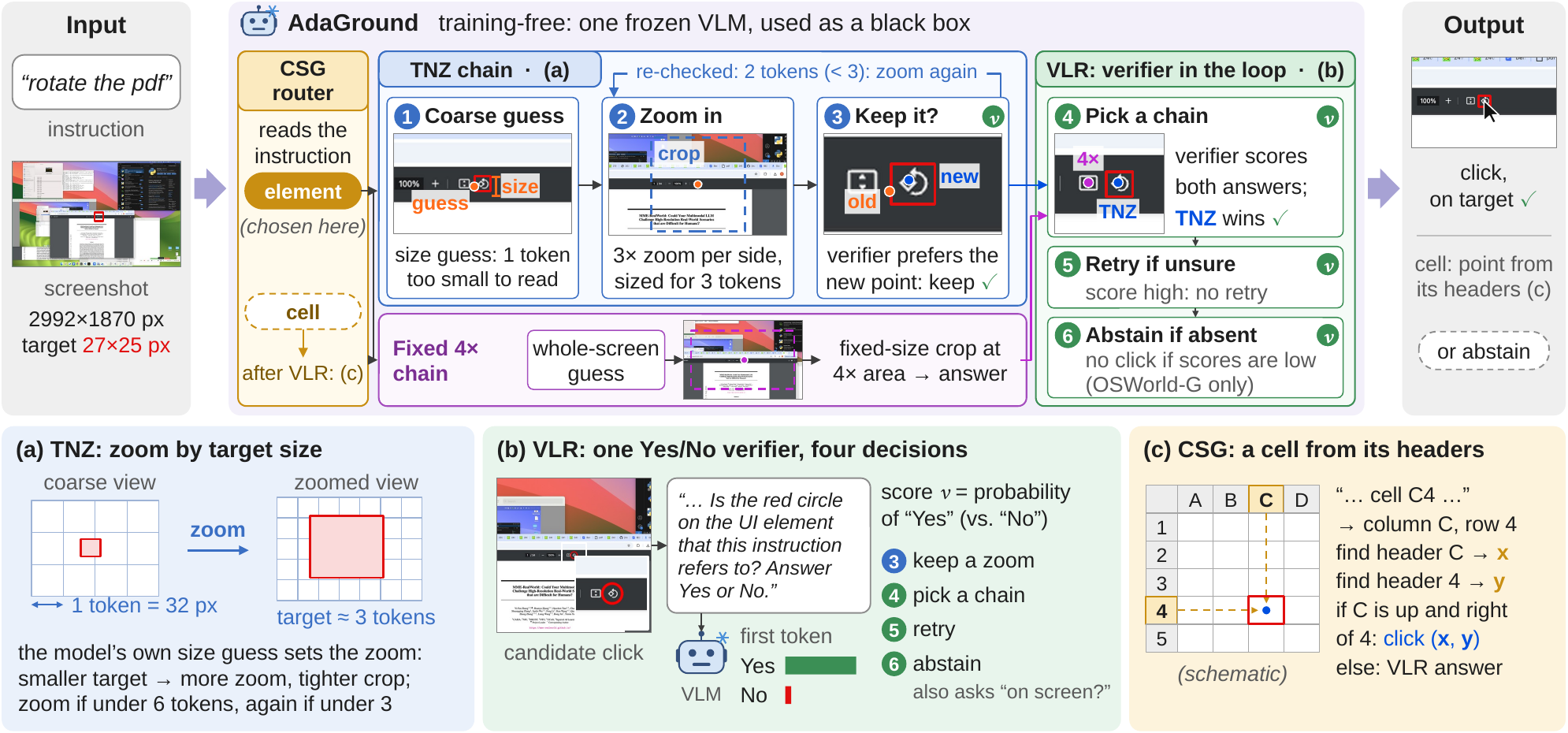}
    \caption{\textbf{Overview of AdaGround.} Compositional Structure Grounding (CSG) reads the instruction. Two zoom chains, Token-Normalized Adaptive Zooming (TNZ) and a fixed $4\times$ crop, run on every sample. Verifier-in-the-Loop Refinement (VLR) uses one Yes/No verifier to keep a zoom, pick a chain, retry with the rejected click crossed out, and abstain. \emph{Top}: the pipeline on one ScreenSpot-Pro sample with Qwen3-VL-8B (real model outputs), where $v$ badges mark the model's Yes/No checks. \emph{Bottom}: (a)~TNZ, (b)~VLR and (c)~CSG, of which (a) and (c) are schematic.}
    \label{fig:hero}
\end{figure*}

\paragraph{Shortcomings of training-free refinement.}
Most current refinement pipelines share five shortcomings.
\textbf{(i)~Scale-agnostic zoom geometry.} Iterative narrowing~\citep{narrowing2024}, ZoomClick~\citep{zoomclick2025}, AdaZoom-GUI~\citep{adazoomgui2026}, UI-Zoomer~\citep{uizoomer2026} and MVP~\citep{mvp2026} set the crop by fixed ratios, size triggers, candidate spread or fixed-size attention windows. None relates the zoom to the number of visual tokens the target will occupy in the rendered view.
\textbf{(ii)~Open-loop acceptance.} The zoomed prediction replaces the coarse one unconditionally, even when the crop has discarded the context that identified the target.
\textbf{(iii)~No second attempt.} Without a check on the answer, a pipeline cannot tell which answers to revisit.
\textbf{(iv)~White-box, vote-based aggregation.} Attention-guided methods~\citep{attngui2025} such as MVP need attention maps from inside the backbone. MVP also aggregates its views by spatial voting, which cannot tell which of two disagreeing views is right.
\textbf{(v)~Always zooming and answering.} Targets absent from the screen are solved by no view, and targets defined by their position in a structure, like a spreadsheet cell, lose their evidence in a tight zoom. Yet every pipeline zooms and answers.

\paragraph{AdaGround.}
Our key observation is that differently zoomed views fail on different samples. On Qwen3-VL-8B ScreenSpot-Pro, a fixed crop and an adaptive one are right on $63.5\%$ and $59.6\%$ of samples, but at least one of them is right on $68.9\%$. Refinement should therefore not commit to one zoom but choose, per sample, which zoomed answer to trust. The grounder itself can make that choice when shown a marked click. We propose \textbf{AdaGround} (Figure~\ref{fig:hero}), a training-free framework in which the grounder sizes its own zoom and verifies its own answers.
\emph{Token-Normalized Adaptive Zooming} (TNZ) answers~(i). From the model's own bounding box on the coarse view, it solves per sample for the magnification, the crop side and the number of zoom stages. They are chosen so that the target spans a fixed number of visual tokens while the crop still covers the coarse localization error.
\emph{Verifier-in-the-Loop Refinement} (VLR) answers (ii)--(iv) and the absent targets of~(v). One marked-click Yes/No verifier drives \emph{verification-gated acceptance}, a \emph{verifier-arbitrated chain portfolio} and \emph{negative-feedback re-grounding}. The portfolio selects between the TNZ chain and a fixed-crop chain by verification instead of by vote, and re-grounding runs with the rejected location crossed out. With a companion existence query, the verifier also decides \emph{existence-gated abstention}.
\emph{Compositional Structure Grounding} (CSG) answers the structural targets of~(v) by composing a spreadsheet cell from its separately grounded column and row headers.
Throughout, the VLM is a black box that takes images and text and returns text and first-token probabilities. One set of constants serves both backbones and all benchmarks (Section~\ref{sec:config}).

\paragraph{Results.}
On OSWorld-G~\citep{osworldg2024} and ScreenSpot-Pro, with Qwen3-VL-8B and Qwen3-VL-32B, AdaGround is above every published result of MVP and ZoomClick (Figure~\ref{fig:teaser}), by up to $7.3$ points. The lead is significant in every paired test against MVP's per-sample predictions. AdaGround is also above all proprietary models and trained GUI specialists in Tables~\ref{tab:osworld_g} and~\ref{tab:main_sspro}, including 72B ones. Its gains are largest on ScreenSpot-Pro's high-resolution screens ($+14.9$ and $+20.6$ points over the single pass) and carry over to held-out UI-Vision~\citep{uivision2025}. On ScreenSpot-v2~\citep{osatlas2024}, whose large targets the single pass already resolves, they are not significant.

\noindent Our contributions are as follows:
\begin{itemize}[leftmargin=*,nosep,topsep=2pt]
\item \textbf{TNZ} sizes every zoom in visual tokens from the model's own estimate of the target, so that the target reaches a legible span at any screen resolution (Property~\ref{prop:resolution}).
\item \textbf{VLR} makes the grounder its own verifier. One marked-click Yes/No question replaces open-loop acceptance and voting, and it decides re-grounding and abstention.
\item \textbf{CSG} composes structural targets from separately grounded headers instead of zooming.
\end{itemize}

\section{Related Work}\label{sec:related_work}
\label{sec:related}

\paragraph{GUI grounding.}
GUI grounding is commonly cast as coordinate generation by a VLM~\citep{seeclick2024,qwen3vl2025,seedvl2025}, following UI-specialised models~\citep{screenai2024,ferretui2024}. Specialists fine-tuned on GUI corpora~\citep{uground2025,osatlas2024,cogagent2024,showui2025,guiactor2025} and reinforcement post-training or self-distillation~\citep{infigui2025,guig2_2025,guig1_2025,vista2026,guisd2026} improve the grounder itself. AdaGround instead changes how an off-the-shelf grounder is queried.

\paragraph{Test-time zooming.}
Iterative narrowing~\citep{narrowing2024} re-predicts on crops of fixed geometry around the previous answer, RegionFocus~\citep{regionfocus2025} and DiMo-GUI~\citep{dimogui2025} zoom into candidate regions, and V$^{\ast}$~\citep{vstar2024} searches for small objects outside GUIs. Three recent systems refine the zoom. ZoomClick~\citep{zoomclick2025} crops with a fixed shrink ratio for a fixed three iterations. AdaZoom-GUI~\citep{adazoomgui2026} uses a trained grounder's box as a small-or-large trigger and then magnifies by a predefined ratio. UI-Zoomer~\citep{uizoomer2026} sizes one crop from the variance of eight sampled candidates, scaled by a model- and benchmark-specific constant. None derives the zoom from the target's token span, and all three accept the zoomed prediction unconditionally (shortcomings~(i) and~(ii)). Other training-free pipelines progressively zoom onto interface elements~\citep{zoomui2026}, refine through multi-step visual reasoning~\citep{chainofground2025} or search over screen regions~\citep{drsgui2026}. Others mitigate grounding biases~\citep{bami2026}, crop coarse-to-fine with OCR and detected components as cues~\citep{guilens2026}, or guide a visual search by the perplexity of coordinate tokens~\citep{autofocus2026}.

\paragraph{Attention-guided and vote-based aggregation.}
MVP~\citep{mvp2026}, the attention-guided zooming method we compare against, extracts attention-cropped views and aggregates their predictions by spatial clustering. Attention-driven grounding~\citep{attngui2025} likewise reads the model's attention, and GUI-RC~\citep{guirc2025} votes over sampled predictions on a spatial grid. Concurrent with our work, Re-Prefill~\citep{lin2026reprefill} selects visual tokens from the attention of a first prefill and modifies the model's prefill in a second pass. Like MVP, it needs model internals. AdaGround reads only the generated text and first-token probabilities, and replaces voting with verification (shortcoming~(iv)).

\paragraph{Verification and feedback.}
GUI-Cursor~\citep{guicursor2025} and UI-Ins~\citep{uiins2025} train spatial feedback and instruction reasoning into the grounder. LookAgain~\citep{lookagain2026} and Measure Twice, Click Once~\citep{measuretwice2026} train the grounder or a visual critic to re-examine its clicks. SafeGround~\citep{safeground2026} and \citet{argus2026} calibrate and benchmark uncertainty in GUI grounding. AdaGround obtains these signals from the unmodified model through one Yes/No question about a marked click and an existence question (shortcomings~(iii) and~(v)).

\paragraph{Learned focusing and agent memory.}
FocusUI~\citep{focusui2026} learns a position-preserving selection of visual tokens, GUI-Spotlight~\citep{guispotlight2025} trains a model to narrow its focus, and DeepEyes~\citep{deepeyes2025} and AdaptVision~\citep{adaptvision2025} learn to request zooms. Agent frameworks add visual procedural memory~\citep{skilllens2026} or evolve skills from traces~\citep{skillevo2026,omegause2026}. AdaGround is training-free and solves whether and how far to zoom from the model's own size estimate.

\section{The AdaGround Framework}
\label{sec:method}

As shown in Figure~\ref{fig:hero} (Algorithm~\ref{alg:adaground} in Appendix~\ref{sec:app_impl}), AdaGround maps a screenshot--instruction pair $(S,I)$ to a click point or an abstention. CSG (Section~\ref{sec:compose}) first reads the instruction. Every sample then runs through two zoom chains, the TNZ chain (steps~1--3 of Figure~\ref{fig:hero}, Section~\ref{sec:zoom}) and the fixed $4\times$ chain. VLR (Section~\ref{sec:verify}) uses one Yes/No verifier to keep a zoom (step~3), pick a chain (step~4), retry (step~5) and abstain (step~6). Finally, a cell flagged by CSG is composed from its headers. All three components share one set of constants (Section~\ref{sec:config}).

\subsection{Problem Formulation}\label{sec:problem}

Given a screenshot $S\in\mathbb{R}^{H\times W\times 3}$ and an instruction $I$, GUI grounding predicts a point $p=(x,y)$, which is correct iff $p\in B$, the ground-truth region of the referenced element. When the instruction has no target on screen, the only correct output is an abstention, and an abstention on any other sample is an error. An off-the-shelf VLM $M$, used without any weight update, receives a rendered view and the instruction and emits a coordinate as text, $p=M(\mathcal{R}(S,b),I)$. The renderer $\mathcal{R}(S,b)$ scales the area of $S$ by $b$, so the linear scale is $\sqrt b$ and $b>1$ up-samples. Access to $M$ is black-box. We read its generated text and first-token probabilities, and never its attention maps, hidden states or gradients.

\paragraph{Token pitch and token span.} Encoders that work at dynamic resolution tile a rendered view into patches and merge neighbouring patches into one visual token. Qwen3-VL~\citep{qwen3vl2025} merges $2\times2$ patches of $16$ pixels, so one token covers a square of side $\pi=32$ rendered pixels. For a target whose shorter side measures $s$ original pixels, shown in a view of linear scale $k$ relative to the screenshot, the \emph{token span}
\begin{equation}
    \sigma \;=\; \frac{s\,k}{\pi}
    \label{eq:span}
\end{equation}
is the number of visual tokens the target occupies along its shorter side in the view the model is given. Because it is measured after rendering, it is comparable across resolutions, crops and up-sampling. Accuracy climbs steeply below about three tokens and gains much less above (Appendix~\ref{sec:app_tokenspan}). This motivates two design choices of TNZ. Every zoom is normalized to a target span of $\tau$ tokens, and the first zoom fires under a looser criterion than later ones.

\subsection{Token-Normalized Adaptive Zooming}\label{sec:zoom}
TNZ decides per sample whether to zoom, how much, how wide and how often. All four decisions follow from one quantity that the model reports, the extent of the target.

\paragraph{Scale-aware coarse localization (step~1).} The screenshot is first rendered at the budget $b_1=\min\bigl(1,\max(b_{\min},A_0/(HW))\bigr)$, which keeps the coarse view near $A_0$ pixels. A 4K screenshot is therefore down-sampled, while a 1080p one is not. The model returns a coarse point $p_1=M(\mathcal{R}(S,b_1),I)$. In a second, separate query on the same view, it also returns the target's bounding box. The shorter side of that box, mapped back to original pixels, is the \emph{self-reported extent} $\hat s$. Let $k_{\text{prev}}=\sqrt{b_1}$ be the linear scale of the view that the current point came from. The estimated span is then $\hat\sigma=\hat s\,k_{\text{prev}}/\pi$. If no box is returned, no zoom is issued. $\hat s$ only sets a scale and need only be right to within a factor (Appendix~\ref{sec:shat_quality}).

\paragraph{Footprint-matched magnification and error-aware field of view (step~2).} TNZ magnifies the crop so that the estimated footprint is rendered at $\tau$ tokens, and sizes it to cover both the target's context and the coarse localization error:
\begin{align}
    k &= \operatorname{clip}\bigl(\tau\pi/\hat s,\ 1,\ k_{\max}\bigr), \label{eq:k}\\
    d &= \operatorname{clip}\bigl(\max(\gamma\hat s,\,\varepsilon\pi/k_{\text{prev}}),\ d_{\min},\ d_{\max}\bigr), \label{eq:adaptive_d}
\end{align}
with $d_{\max}=\min(W,H)$. The term $\gamma\hat s$ keeps the context proportional to the target. The term $\varepsilon\pi/k_{\text{prev}}$ reserves $\varepsilon$ tokens of the view the current point came from, where one token covers $\pi/k_{\text{prev}}$ original pixels. The square crop of side $d$ is centred on the current point and shifted inside the screenshot.
\begin{property}[Invariance and coverage]\label{prop:resolution}\label{prop:coverage}
(a)~If the estimate is exact, $\hat s=s$, and $\tau\pi/k_{\max}\le s\le\tau\pi$, the zoomed view renders the target at $\sigma=\tau$ tokens whatever the screenshot resolution (up to the renderer's rounding). (b)~If the current point is off the target centre by at most $\varepsilon/2$ tokens of the previous view along each axis, the target centre lies inside the crop (provided $\varepsilon\pi/k_{\text{prev}}\le d_{\max}$).
\end{property}
\noindent\emph{Argument.} (a)~On this range Eq.~\eqref{eq:k} is unclipped, so Eq.~\eqref{eq:span} gives $\sigma'=s\,k/\pi=\tau$, in which neither $H$, $W$ nor $b_1$ appears. (b)~$d/2\ge\varepsilon\pi/(2k_{\text{prev}})$, which is $\varepsilon/2$ tokens of the previous view in original pixels. Shifting the box inside the screenshot keeps the in-screen part of the centred box.\hfill$\square$

\noindent With Qwen3-VL and the constants of Section~\ref{sec:config} the range of (a) is $24$ to $96$ pixels. Below this range, $k$ saturates at $k_{\max}$. Above it, $k=1$ gives a native-resolution crop, which gated acceptance may reject.

\paragraph{Magnified re-grounding.} The crop $C$, whose actual top-left corner after shifting is $(x_0,y_0)$, is rendered at area $k^2$ and re-grounded. The answer, given in pixels of the zoomed view, is mapped back to the screenshot:
\begin{equation}
    p' \;=\; (x_0,y_0) + M\bigl(\mathcal{R}(C,k^2),\,I\bigr)/k.
    \label{eq:map_back}
\end{equation}

\paragraph{Adaptive depth.} Each stage re-crops the screenshot around the current answer. Stage $t$ fires iff $\hat\sigma<\lambda_t\tau$ and, for $t\ge2$, $k_{\text{prev}}<k_{\max}$, where $\lambda_1=\lambda>1$ and $\lambda_t=1$ for $t\ge2$. At most $D$ stages run. After every stage but the last, the box query is repeated on the zoomed view. The new box updates $\hat s$ (in original pixels) whether or not $p'$ is accepted, while $k_{\text{prev}}\leftarrow k$ only when $p'$ is accepted (Eq.~\eqref{eq:accept}). The depth, from zero to $D$ stages, is thus decided per sample. A target whose coarse estimate already spans $\lambda\tau$ tokens gets no crop.

\begin{table*}[t]
\centering
\footnotesize
\setlength{\tabcolsep}{4pt}
\renewcommand{\arraystretch}{0.9}
\begin{tabular}{@{}lcccccc@{}}
\toprule
\textbf{Method} & \textbf{Text} & \textbf{Element} & \textbf{Layout} & \textbf{Fine-gr.} & \textbf{Refusal} & \textbf{Overall} \\
\midrule
\rowcolor{gray!15} \multicolumn{7}{l}{\textit{Closed-source / proprietary models}} \\
Operator\smash{${}^{\dagger\circ}$}~\citep{operator2025} & 51.3 & 42.4 & 46.6 & 31.5 & 0.0 & 40.6 \\
Gemini-2.5-pro\smash{${}^{\dagger\circ}$}~\citep{gemini2025} & 59.8 & 45.5 & 49.0 & 33.6 & 38.9 & 45.2 \\
Seed1.5-VL\smash{${}^{\dagger\circ}$}~\citep{seedvl2025} & 73.9 & 66.7 & 69.6 & 47.0 & 18.5 & 62.9 \\
\midrule
\rowcolor{gray!15} \multicolumn{7}{l}{\textit{Open-source SFT / RL-trained specialists}} \\
Qwen2.5-VL-72B\smash{${}^{\dagger}$}~\citep{qwen25vl2025} & 52.6 & 74.6 & 74.7 & 55.3 & 0.0 & 62.2 \\
UI-TARS-72B\smash{${}^{\dagger\circ}$}~\citep{uitars2025} & 69.4 & 60.6 & 62.9 & 45.6 & 0.0 & 57.1 \\
UI-Venus-Ground-72B\smash{${}^{\dagger}$}~\citep{uivenus2025} & 82.1 & 71.2 & 70.7 & 64.4 & -- & 70.4 \\
OS-Atlas-7B\smash{${}^{\dagger\circ}$}~\citep{osatlas2024} & 44.1 & 29.4 & 35.2 & 16.8 & 7.4 & 27.7 \\
UGround-V1-7B\smash{${}^{\dagger\circ}$}~\citep{uground2025} & 51.3 & 40.3 & 43.5 & 24.8 & 0.0 & 36.4 \\
Jedi-7B\smash{${}^{\dagger\circ}$}~\citep{osworldg2024} & 65.9 & 55.5 & 57.7 & 46.9 & 7.4 & 54.1 \\
GTA1-72B\smash{${}^{\dagger}$}~\citep{gta1_2025} & 57.9 & 76.9 & 77.3 & 66.7 & 0.0 & 66.7 \\
\midrule
\rowcolor{gray!15} \multicolumn{7}{l}{\textit{Training-free test-time methods}} \\
Qwen3-VL-8B single pass & 78.2 & 68.5 & 71.9 & 53.7 & 0.0 & 64.7 \\
\quad + MVP\smash{${}^{\dagger}$} (published, from its $68.8$ single pass) & 82.4 & 76.7 & 79.5 & 61.7 & -- & 72.7\gain{3.9} \\
\rowcolor{table_color!20} \quad \textbf{+ AdaGround (ours)} & 85.1 & 80.0 & 81.8 & 65.1 & \textbf{27.8} & 76.1\gain{11.3} \\
\midrule
Qwen3-VL-32B\smash{$^{\ast}$} single pass & 81.6 & 75.8 & 77.5 & 63.1 & 0.0 & 71.8 \\
\quad + MVP\smash{${}^{\dagger}$} (published, from its $71.7$ single pass) & 81.6 & 76.4 & 77.5 & 62.4 & -- & 72.0\gain{0.3} \\
\rowcolor{table_color!20} \quad \textbf{+ AdaGround (ours)} & \textbf{85.8} & \textbf{84.8} & \textbf{84.6} & \textbf{67.1} & 14.8 & \textbf{79.3}\gain{7.4} \\
\bottomrule
\end{tabular}
\caption{\textbf{OSWorld-G} accuracy (\%). Refusal: abstention rate on the $54$ absent-target instructions. ${}^{\dagger}$Quoted from the cited papers or the tables of \citet{gta1_2025} and \citet{mvp2026}. ${}^{\circ}$Original instructions (Appendix~\ref{sec:statistics}). --: not reported. \textbf{Bold}: best training-free result. {\color{gaincolor}Subscripts}: gain over the method's own single pass. $^{\ast}$FP8 weights.}
\label{tab:osworld_g}
\end{table*}

\subsection{Verifier-in-the-Loop Refinement}\label{sec:verify}
TNZ proposes and VLR decides. A black-box Yes/No verifier and a companion existence query close the loop around four decisions that earlier pipelines take open-loop or through model internals.

\paragraph{The verifier.} For a candidate point $p$,
\begin{equation}
    v(p) \;=\; \frac{P_M(\textsc{Yes}\mid x_p)}{P_M(\textsc{Yes}\mid x_p)+P_M(\textsc{No}\mid x_p)},
    \label{eq:verify}
\end{equation}
where $x_p=(\mathcal{V}(S,p),q(I))$ and first-token probabilities are summed over the case variants of each word. The verification view $\mathcal{V}(S,p)$ is the window of side $d_0(W,H)=\operatorname{clip}(\lfloor\sqrt{WH/8}\rfloor,512,1024)$ centred on $p$ and shifted inside the screenshot. It is rendered at $4\times$ area with a red circle drawn at $p$. The question $q(I)$ asks whether the circle lies on the element the instruction refers to (Appendix~\ref{sec:prompts_verbatim}).

\paragraph{Verification-gated acceptance (step~3).} A TNZ stage's answer $p'$ replaces the current answer $p$ only if the verifier does not prefer $p$,
\begin{equation}
    p \;\leftarrow\; p' \quad\text{iff}\quad v(p')\ge v(p),
    \label{eq:accept}
\end{equation}
and $v(p)$ is queried only when a zoom is planned, so unzoomed samples cost no verification.

\paragraph{Verifier-arbitrated chain portfolio (step~4).} Chains of different zoom geometry fail on different samples. On Qwen3-VL-8B ScreenSpot-Pro, the chains $c_{\mathrm{TNZ}}$ and $c_{4\times}$ defined below are right on $59.6\%$ and $63.5\%$ of samples, but at least one of them is right on $68.9\%$ (Appendix~\ref{sec:extended_ablation}). VLR therefore runs a static portfolio $\mathcal{C}=\{c_{\mathrm{TNZ}},c_{4\times}\}$, the same for every input, benchmark and backbone.
Here $c_{\mathrm{TNZ}}$ is TNZ with gated acceptance, and $c_{4\times}$ runs the common fixed $4\times$ crop as a chain. It first answers on the whole screen and then re-grounds in a $1288\times728$-pixel window around that answer at $4\times$ area. The answers $\{p^{(c)}\}_{c\in\mathcal{C}}$ are grouped by complete linkage with radius $\delta_{\text{px}}$, so a point joins a group only if it lies within $\delta_{\text{px}}$ of every member. Each group $G$ is represented by its member $r(G)$ from the highest-priority chain, with $c_{\mathrm{TNZ}}$ first. With two or more groups, the representatives of the $K$ largest groups, $\mathcal{G}_K$, are verified, each on its own window, and
\begin{equation}
    p \;=\; \operatorname*{arg\,max}_{G\in\mathcal{G}_K}\; v\bigl(r(G)\bigr);
    \label{eq:arbitrate}
\end{equation}
with a single group, its representative is returned. With the two chains used here, arbitration thus keeps the better-verified of two disagreeing answers, whereas the spatial vote of MVP~\citep{mvp2026} keeps the largest cluster.

\paragraph{Negative-feedback re-grounding (step~5).} When the verifier does not confirm the answer kept by arbitration, $v(p)<\theta_r$, the model is asked again, but on different evidence. The rejected location is crossed out in red on the whole-screen view, and a note saying that it was judged not to be the target is appended to the instruction for that pass. The model then re-grounds from coarse to fine. It answers on the whole screenshot, then on a native-resolution crop of side $d_0$ around the new point, and finally on a tighter crop of side $d_0/2$ around the resulting point at $3\times$ linear magnification, which gives $p''$. The answer is replaced, $p\leftarrow p''$, iff $v(p'')>v(p)$.

\paragraph{Existence-gated abstention (step~6).} Let $e(S,I)$ be the normalized first-token \textsc{Yes} probability, as in Eq.~\eqref{eq:verify}, of one query on the whole screenshot ($b=1$) asking whether any element on the screen can carry out the instruction. Neither signal alone is a safe reason to refuse. The existence query denies targets that are present but tiny on the whole-screen view, and the verifier rejects answers it cannot confirm at its own scale. AdaGround therefore abstains only when the mean of the two is low, which requires each to be below $2\theta_e$,
\begin{equation}
    \tfrac12\bigl(v(p)+e(S,I)\bigr) \;<\; \theta_e,
    \label{eq:abstain}
\end{equation}
where $p$ is the answer after re-grounding, so $v(p)$ is the larger of the scores before and after re-grounding. An answer the verifier accepts ($v(p)\ge\theta_r$) is never withdrawn, and abstention never overrides a CSG-composed answer. Abstention is enabled only where the task admits a refusal (OSWorld-G).

\subsection{Compositional Structure Grounding}\label{sec:compose}
A spreadsheet cell is defined by its place in a structure, not by its appearance. The cell is identified by its column and row headers. These lie outside any crop tight enough to magnify the cell, so zooming tends to discard the evidence. CSG changes the primitive instead of the view. The same model, reading the instruction text only, rewrites $I$ into a program $\rho\in\{\textsc{ground},\textsc{cell}(\mathit{col},\mathit{row})\}$ (the router of Figure~\ref{fig:hero}, whose `element' branch is \textsc{ground}). Any other output is executed as \textsc{ground}. \textsc{cell}$(\mathit{col},\mathit{row})$ grounds descriptions of the column header and of the row header separately. Each header is grounded first on the whole screenshot and then on a crop of side $d_0/2$ around that point at $3\times$ linear magnification, which gives $(x_{\mathrm{col}},y_{\mathrm{col}})$ and $(x_{\mathrm{row}},y_{\mathrm{row}})$. The program then composes
\begin{equation}
    p=(x_{\mathrm{col}},y_{\mathrm{row}})\ \text{iff}\ x_{\mathrm{col}}>x_{\mathrm{row}},\,y_{\mathrm{col}}<y_{\mathrm{row}},
    \label{eq:cell}
\end{equation}
i.e.\ when the column header lies to the right of and above the row header, as in a spreadsheet grid. Otherwise it falls back to the VLR answer.

\subsection{One Configuration}\label{sec:config}
One set of constants serves both backbones and all benchmarks (Table~\ref{tab:constants}). The coarse view uses $A_0=2.2$\,MP and $b_{\min}=0.25$. TNZ uses $\tau=3$, $\gamma=16$, $\varepsilon=4$, $k_{\max}=4$, $\lambda=2$, $d_{\min}=192$ and $D=2$. The verifier window has side $d_0$ and is rendered at $4\times$ area, and $c_{4\times}$ uses a $1288\times728$ window at $4\times$ area. Arbitration uses $\delta_{\text{px}}=14$ and $K=3$, re-grounding uses $\theta_r=0.99$ (whole screen at $b=1$), and abstention uses $\theta_e=0.03$. Only $\pi$ comes from the encoder.

\section{Experiments}\label{sec:exp}
\begin{table*}[t]
\centering
\footnotesize
\setlength{\tabcolsep}{4pt}
\renewcommand{\arraystretch}{0.9}
\begin{tabular}{@{}lccccccc@{}}
\toprule
\textbf{Method} & \textbf{Dev.} & \textbf{Creative} & \textbf{CAD} & \textbf{Scientific} & \textbf{Office} & \textbf{OS} & \textbf{Overall} \\
\midrule
\rowcolor{gray!15} \multicolumn{8}{l}{\textit{Closed-source / proprietary models}} \\
GPT-4o\smash{${}^{\dagger}$}~\citep{gpt4o2024} & 0.7 & 0.6 & 1.5 & 1.2 & 0.9 & 0.0 & 0.8 \\
Claude Computer Use\smash{${}^{\dagger}$}~\citep{claudecomputeruse2024} & 12.6 & 16.8 & 11.9 & 25.8 & 26.9 & 8.1 & 17.1 \\
UI-TARS-1.5\smash{${}^{\dagger}$}~\citep{uitars2025} & 63.9 & 50.4 & 58.2 & 69.3 & 79.6 & 51.0 & 61.6 \\
Seed1.5-VL\smash{${}^{\dagger}$}~\citep{seedvl2025} & 53.8 & 59.2 & 59.0 & 61.4 & 74.8 & 60.2 & 60.9 \\
\midrule
\rowcolor{gray!15} \multicolumn{8}{l}{\textit{Open-source SFT / RL-trained specialists}} \\
SeeClick-7B\smash{${}^{\dagger}$}~\citep{seeclick2024} & 0.3 & 0.6 & 1.9 & 2.0 & 0.9 & 1.5 & 1.1 \\
UGround-V1-7B\smash{${}^{\dagger}$}~\citep{uground2025} & 28.1 & 31.7 & 14.6 & 39.0 & 49.6 & 24.5 & 31.1 \\
GUI-G1-3B\smash{${}^{\dagger}$}~\citep{guig1_2025} & 31.1 & 26.6 & 32.2 & 48.0 & 59.1 & 16.1 & 37.1 \\
SE-GUI-7B\smash{${}^{\dagger}$}~\citep{segui2025} & 44.5 & 37.2 & 42.1 & 54.7 & 70.4 & 38.8 & 47.2 \\
InfiGUI-G1-7B\smash{${}^{\dagger\S}$}~\citep{infigui2025} & 50.2 & 44.0 & 49.0 & 59.4 & 67.4 & 44.4 & 51.9 \\
GTA1-72B\smash{${}^{\dagger}$}~\citep{gta1_2025} & 57.2 & 51.0 & 49.8 & 63.0 & 77.0 & 57.1 & 58.4 \\
UI-Venus-Ground-72B\smash{${}^{\dagger\S}$}~\citep{uivenus2025} & 59.5 & 55.4 & 57.5 & 66.5 & 77.8 & 57.7 & 61.9 \\
\midrule
\rowcolor{gray!15} \multicolumn{8}{l}{\textit{Training-free test-time methods}} \\
Qwen3-VL-8B single pass & 47.2 & 46.0 & 44.4 & 53.9 & 71.3 & 49.0 & 51.3 \\
\quad + MVP\smash{${}^{\dagger}$} (published, from its $55.0$ single pass) & 61.5 & 60.2 & 61.3 & 67.3 & 82.6 & 62.8 & 65.3\gain{10.3} \\
\rowcolor{table_color!20} \quad \textbf{+ AdaGround (ours)} & 61.5 & 57.2 & 65.5 & 66.9 & 85.2 & 66.3 & 66.2\gain{14.9} \\
\midrule
Qwen3-VL-32B\smash{$^{\ast}$} single pass & 43.1 & 53.7 & 57.1 & 63.0 & 73.0 & 43.9 & 55.3 \\
\quad + MVP\smash{${}^{\dagger}$} (published, from its $55.3$ single pass) & 71.6 & 69.3 & 74.7 & 70.5 & 87.4 & 73.5 & 74.0\gain{18.7} \\
\quad + ZoomClick\smash{${}^{\dagger}$}~\citep[from its $54.0$ single pass;][]{zoomclick2025} & \textbf{72.6} & 64.2 & 71.3 & 68.9 & 90.0 & 69.4 & 72.1\gain{18.1} \\
\rowcolor{table_color!20} \quad \textbf{+ AdaGround (ours)} & 71.6 & \textbf{70.1} & \textbf{76.6} & \textbf{74.8} & \textbf{91.3} & \textbf{75.0} & \textbf{75.9}\gain{20.6} \\
\bottomrule
\end{tabular}
\caption{\textbf{ScreenSpot-Pro} accuracy (\%). ${}^{\dagger}$Quoted from the cited papers or as compiled by \citet{gta1_2025} and \citet{mvp2026}. ${}^{\S}$Category averages computed from the reported text/icon accuracies. \textbf{Bold}: best training-free result. {\color{gaincolor}Subscripts}: gain over the method's own single pass. $^{\ast}$FP8 weights.}
\label{tab:main_sspro}
\end{table*}

\begin{table}[t]
\centering
\footnotesize
\setlength{\tabcolsep}{3pt}
\renewcommand{\arraystretch}{0.95}
\resizebox{\columnwidth}{!}{\begin{tabular}{@{}lcccc@{}}
\toprule
\textbf{Method} & \textbf{Basic} & \textbf{Func.} & \textbf{Spatial} & \textbf{Avg.} \\
\midrule
\rowcolor{gray!15} \multicolumn{5}{l}{\textit{Open-source SFT / RL-trained specialists}${}^{\dagger}$} \\
UI-TARS-7B~\citep{uitars2025} & 20.1 & 24.3 & 8.4 & 17.6 \\
UI-TARS-72B~\citep{uitars2025} & 31.4 & 30.5 & 14.7 & 25.5 \\
UI-Venus-Ground-72B~\citep{uivenus2025} & 45.6 & 42.3 & 23.7 & 36.8 \\
OS-Atlas-7B~\citep{osatlas2024} & 12.2 & 11.2 & 3.7 & 9.0 \\
UGround-V1-7B~\citep{uground2025} & 15.4 & 17.1 & 6.3 & 12.9 \\
GUI-Spotlight~\citep{guispotlight2025} & 32.1 & 30.2 & 9.1 & 23.4 \\
\midrule
\rowcolor{gray!15} \multicolumn{5}{l}{\textit{Training-free test-time methods}} \\
Qwen3-VL-8B single pass & 29.2 & 30.1 & 11.8 & 23.7 \\
\quad + MVP\smash{${}^{\dagger}$} & 38.3 & 38.8 & 18.7 & 31.9\gain{4.7} \\
\rowcolor{table_color!20} \quad \textbf{+ AdaGround (ours)} & 37.0 & 38.7 & 20.4 & 32.0\gain{8.3} \\
\midrule
Qwen3-VL-32B\smash{$^{\ast}$} single pass & 44.2 & 43.2 & 22.6 & 36.7 \\
\quad + MVP\smash{${}^{\dagger}$} & 49.4 & \textbf{52.0} & 30.8 & 44.1\gain{7.7} \\
\quad + ZoomClick\smash{${}^{\dagger}$} & 45.1 & 45.1 & 28.3 & 39.2\gain{2.6} \\
\rowcolor{table_color!20} \quad \textbf{+ AdaGround (ours)} & \textbf{52.5} & 51.7 & \textbf{38.4} & \textbf{47.6}\gain{10.9} \\
\bottomrule
\end{tabular}}
\caption{\textbf{UI-Vision} element grounding~\citep{uivision2025} (held out), accuracy (\%). Avg.: mean of the three tasks (as published for ${}^{\dagger}$ rows). ${}^{\dagger}$From \citet{mvp2026} (specialists from the UI-Vision leaderboard) and \citet{zoomclick2025}. {\color{gaincolor}Subscripts}: gain over the method's own single pass (MVP: 27.2, 36.4; ZoomClick: 36.6). \textbf{Bold}: best training-free. $^{\ast}$FP8 weights.}
\label{tab:uivision}
\end{table}

\begin{table}[t]
\centering
\footnotesize
\setlength{\tabcolsep}{2.2pt}
\renewcommand{\arraystretch}{0.95}
\begin{tabular}{@{}lcccccc@{}}
\toprule
 & \multicolumn{2}{c}{\textbf{OSWorld-G}} & \multicolumn{2}{c}{\textbf{SSPro}} & \multicolumn{2}{c}{\textbf{SSv2}} \\
\cmidrule(lr){2-3}\cmidrule(lr){4-5}\cmidrule(l){6-7}
 & 8B & 32B$^{\ast}$ & 8B & 32B$^{\ast}$ & 8B & 32B$^{\ast}$ \\
\midrule
Single pass & 64.7 & 71.8 & 51.3 & 55.3 & 92.1 & 94.7 \\
Fixed $4\times$ crop & 70.7 & \underline{77.1} & \underline{63.5} & \underline{73.9} & 91.9 & \underline{95.4} \\
\midrule
$c_{\mathrm{TNZ}}$ & \underline{72.7} & 76.6 & 59.6 & 72.4 & \textbf{\underline{92.5}} & 95.4 \\
\quad + arbitration & 73.6 & 76.8 & 65.6 & 75.6 & \textbf{92.5} & \textbf{96.0} \\
\quad + re-grounding & 73.2 & 77.0 & \textbf{66.2} & \textbf{76.0} & 91.9 & 95.4 \\
\quad + CSG & 73.6 & 77.8 & \textbf{66.2} & 75.9 & 91.9 & 95.4 \\
\rowcolor{table_color!20} \quad + abstention & \textbf{76.1} & \textbf{79.3} & \textbf{66.2} & 75.9 & 91.9 & 95.4 \\
\midrule
\textit{Oracle of two chains} & \textit{76.1} & \textit{79.8} & \textit{68.9} & \textit{78.8} & \textit{94.0} & \textit{96.6} \\
\bottomrule
\end{tabular}
\caption{\textbf{Where the gains come from}: accuracy (\%) as components are added (Algorithm~\ref{alg:adaground}). Shaded: AdaGround. \textbf{Bold}: best per column (oracle excluded). \underline{Underlined}: the better chain. Oracle: either chain is right. SSPro/SSv2: ScreenSpot-Pro/-v2. $^{\ast}$FP8 weights.}
\label{tab:ladder}
\end{table}

\subsection{Experimental Setup}
\label{sec:setup}

\paragraph{Backbones and benchmarks.}
Following MVP \citep{mvp2026}, we evaluate Qwen3-VL-8B-Instruct and Qwen3-VL-32B-Instruct~\citep{qwen3vl2025}, the latter with FP8 weights. We develop on three benchmarks, each in full. \textbf{OSWorld-G}~\citep{osworldg2024} (refined split, $564$ instructions) includes $54$ instructions whose target is absent from the screen. \textbf{ScreenSpot-Pro}~\citep{screenspotpro2025} ($1{,}581$ instructions) covers professional desktop software on high-resolution screens, with a median target of $24$ pixels on its shorter side. \textbf{ScreenSpot-v2}~\citep{osatlas2024} ($1{,}272$ instructions) has larger targets (median $85$ pixels). As a held-out test we add \textbf{UI-Vision}~\citep{uivision2025} (the element grounding task, with $5{,}479$ instructions on $1{,}181$ screenshots of $83$ desktop applications and a median target of $26$ pixels).

\paragraph{Baselines.}
The \emph{single pass} grounds on the whole screenshot and comes from the same runs as AdaGround. \emph{MVP}~\citep{mvp2026}, a recent attention-guided zooming method, is our main comparison on the same backbones. \emph{ZoomClick}~\citep{zoomclick2025} reports Qwen3-VL-32B results. The main tables quote both as published. For paired tests, we also reproduce MVP with its official code and published configuration on Qwen3-VL-8B. For Qwen3-VL-32B, we score the per-sample ScreenSpot-Pro predictions released by its authors (Appendices~\ref{sec:overall} and~\ref{sec:efficiency_app}). The fixed $4\times$ crop, AdaGround's chain $c_{4\times}$ alone, is analysed as an ablation (Section~\ref{sec:components}).

\paragraph{Protocol.}
A prediction is correct if its point falls inside the target region. OSWorld-G uses its official scorer, which credits an absent-target instruction only when the system abstains. ScreenSpot-Pro, ScreenSpot-v2 and UI-Vision give every instruction a target and are scored point-in-box. AdaGround and its baselines use greedy decoding on vLLM~$0.11.2$~\citep{vllm2023}. All constants were fixed during development on OSWorld-G, ScreenSpot-Pro and ScreenSpot-v2 (Appendix~\ref{sec:app_impl}). UI-Vision, on which the configuration was neither developed nor tuned, serves as a held-out test. An earlier version of the method had been evaluated on it (Appendix~\ref{sec:app_impl}). Paired differences are in points, with $95\%$ confidence intervals from $4{,}000$ bootstrap resamples of screenshots (Table~\ref{tab:paired}, Appendix~\ref{sec:statistics}). A difference is called significant when its interval excludes zero.

\subsection{Main Results}
\label{sec:main_results}

\paragraph{Comparison with MVP and ZoomClick.}
AdaGround scores above MVP's published results on OSWorld-G and ScreenSpot-Pro with both backbones (Tables~\ref{tab:osworld_g} and~\ref{tab:main_sspro}), by $3.4$ and $7.3$ points on OSWorld-G and by $0.9$ and $1.9$ on ScreenSpot-Pro, and above ZoomClick's published Qwen3-VL-32B result on ScreenSpot-Pro. Measured from each method's own single pass, AdaGround adds more than MVP and ZoomClick in every case. Paired against MVP's per-sample predictions, the lead is significant in every test (Table~\ref{tab:paired}). It is $+6.2$ and $+4.2$ points on OSWorld-G and ScreenSpot-Pro against our Qwen3-VL-8B reproduction, which recovers the gain MVP adds to its own single pass, and $+1.8$ against MVP's released Qwen3-VL-32B ScreenSpot-Pro predictions. The OSWorld-G lead holds without abstention, which MVP does not use. Our reproduction and its single pass both score below MVP's published results (Appendix~\ref{sec:efficiency_app}).

\paragraph{Held-out benchmark.} On UI-Vision, AdaGround matches MVP's published average with Qwen3-VL-8B ($32.0\%$ against $31.9\%$) and exceeds it with Qwen3-VL-32B ($47.6\%$ against $44.1\%$), as well as ZoomClick's $39.2\%$ (Table~\ref{tab:uivision}).

\subsection{Ablation: Where the Gains Come From}
\label{sec:components}
Table~\ref{tab:ladder} adds the components one at a time (Appendix~\ref{sec:extended_ablation}).
\textbf{Fixed crop.} The full AdaGround is significantly above the fixed $4\times$ crop chain $c_{4\times}$ on ScreenSpot-Pro, on held-out UI-Vision (most on spatial instructions) and on OSWorld-G with Qwen3-VL-8B (Table~\ref{tab:paired}).
\textbf{Token span} explains where a zoom pays. On the same samples before and after a crop pass (Qwen3-VL-8B), accuracy rises by $+7.8$ and $+10.8$ points when the target spans under one token and stops rising from two tokens on (Appendix~\ref{sec:app_tokenspan}), so TNZ targets $\tau=3$ tokens.
\begin{figure}[t]
\centering
\includegraphics[width=\columnwidth]{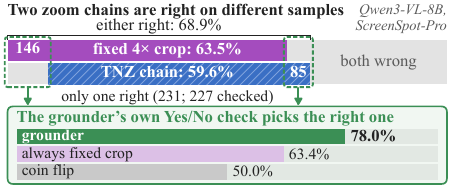}
\caption{\textbf{Chain arbitration} (counts: samples on which only that chain is right). The Yes/No check runs whenever the two clicks differ. With Qwen3-VL-32B the three bars read $72.4$, $56.9$ and $50.0\%$.}
\label{fig:arbitration}
\end{figure}
\textbf{TNZ} as a gated chain ($c_{\mathrm{TNZ}}$) improves the single pass by $4.8$ to $17.1$ points on OSWorld-G and ScreenSpot-Pro and beats the fixed crop on held-out UI-Vision. On ScreenSpot-Pro the two chains disagree on $45$ and $34\%$ of samples, and \textbf{chain arbitration} (Figure~\ref{fig:arbitration}) raises $c_{\mathrm{TNZ}}$ there by $5.9$ and $3.2$ points, $39$ and $36\%$ of the way from the better chain to the oracle. A crop fixed at TNZ's median geometry arbitrates as well on OSWorld-G and ScreenSpot-Pro but significantly worse on ScreenSpot-v2 and held-out UI-Vision (Qwen3-VL-8B, Appendix~\ref{sec:app_controls}).
\textbf{CSG} recovers the OSWorld-G spreadsheet cells that zooming loses ($2$ and $5$ samples net).
\textbf{Abstention} adds $2.5$ and $1.4$ points on OSWorld-G and withdraws at most one correct answer (Table~\ref{tab:gate_ablation}).
\textbf{Cost.} The two chains with arbitration alone (about $10$ calls) already beat the fixed crop on ScreenSpot-Pro (Table~\ref{tab:ladder}).

\section{Conclusion}\label{sec:conclusion}
We presented \textbf{AdaGround}, a training-free, black-box framework that addresses five shortcomings of test-time refinement for GUI grounding. With one configuration shared by Qwen3-VL-8B and Qwen3-VL-32B, AdaGround scores above MVP's published results on OSWorld-G and ScreenSpot-Pro and above many trained GUI models, including 72B ones. It is also significantly above its own fixed $4\times$ crop chain on ScreenSpot-Pro and on held-out UI-Vision. The lesson for test-time refinement is that zoom geometries fail on different samples and that the grounder itself, asked about a marked click, can choose between them. On ScreenSpot-Pro, this choice recovers more than a third of the headroom above the better chain.

\section*{Limitations}
\paragraph{Scope.} We evaluate Qwen3-VL-8B and Qwen3-VL-32B (FP8 weights) on OSWorld-G, ScreenSpot-Pro and ScreenSpot-v2, on which the constants were developed, and on held-out UI-Vision. Our conclusions are limited to these two backbones and four benchmarks and do not extend to agentic tasks in which grounding is one step of a longer episode. On Qwen3-VL-32B OSWorld-G, the comparison with MVP rests on its published number and is not paired.

\paragraph{Cost.} AdaGround reads $3.5$ to $3.9$ times the visual tokens of the fixed $4\times$ crop and $6$ to $12$ times those of the single pass. With sequential calls, its latency is $3.4$ to $4.2$ times that of the fixed crop (Appendix~\ref{sec:app_compute}). The extra compute pays off where targets are small. On ScreenSpot-v2, whose targets are large, AdaGround matches the fixed crop at a higher cost.

\paragraph{Components.} A second crop fixed at TNZ's median geometry is not significantly different from TNZ in arbitration on OSWorld-G and ScreenSpot-Pro, nor in the full pipeline on any benchmark, at fewer visual tokens (Appendix~\ref{sec:app_controls}). TNZ assumes that the model answers the box query with a box. A model that returns a point instead gets no zoom from TNZ. A target that falls outside an over-tight view is rejected by verification and left to the fixed $4\times$ chain and re-grounding.

\paragraph{Verifier and abstention.} One verifier score controls acceptance, re-grounding, arbitration and abstention, so a verifier that prefers a plausible but wrong element passes that preference to every decision it controls. Many of its decisions are between scores near $1$ (Appendix~\ref{sec:extended_ablation}). The joint abstention rule is set for few false refusals ($0.8\%$ and $0.4\%$; Appendix~\ref{sec:mech_validation}) and catches $15$ and $8$ of the $54$ absent-target instructions of OSWorld-G.

\paragraph{Structure.} CSG covers spreadsheet cells but does not handle slider positions or text-cursor placement. On these cells, the single-pass answer does as well as the composed one (Appendix~\ref{sec:extended_ablation}).

\section*{Ethical Considerations}
AdaGround is an inference-time procedure that trains no model and collects no new data. All experiments use the screenshots and instructions of four public GUI grounding benchmarks and involve no human subjects. Better grounding makes GUI agents more capable, including agents that act without supervision. A wrong click can trigger an unintended, possibly irreversible action, and capable agents can be misused to automate interactions that services intend for humans. Abstention removes one failure mode but does not replace user confirmation before consequential actions.

{\setlength{\emergencystretch}{1em}\exhyphenpenalty=10000 \bibliography{references}}

\appendix
\section*{Appendix Contents}
\startcontents[appendix]
\printcontents[appendix]{l}{1}{\setcounter{tocdepth}{2}}

\setcounter{topnumber}{3}\setcounter{dbltopnumber}{3}\setcounter{totalnumber}{5}
\renewcommand{\topfraction}{0.9}\renewcommand{\dbltopfraction}{0.9}\renewcommand{\textfraction}{0.07}
\renewcommand{\floatpagefraction}{0.8}\renewcommand{\dblfloatpagefraction}{0.8}

\section{Implementation Details and Prompts}\label{sec:app_impl}
Table~\ref{tab:constants} lists the constants and Algorithm~\ref{alg:adaground} the procedure.

\begin{table*}[t]
\centering

\footnotesize
\setlength{\tabcolsep}{4pt}
\renewcommand{\arraystretch}{1.05}
\begin{tabular}{@{}>{\raggedright\arraybackslash}p{0.14\linewidth}>{\raggedright\arraybackslash}p{0.36\linewidth}>{\raggedright\arraybackslash}p{0.44\linewidth}@{}}
\toprule
\textbf{Component} & \textbf{Constant} & \textbf{Value} \\
\midrule
Rendering & grid; pixels per view & $32$\,px; $3{,}136$ to $8{,}847{,}360$ \\
\midrule
Coarse view & pixel budget $A_0$; floor $b_{\min}$ of $b_1$ & $2.2$\,MP; $0.25$ \\
\midrule
TNZ & token pitch $\pi$ & $32$ (Qwen3-VL) \\
 & target span $\tau$ & $3$ tokens \\
 & context multiple $\gamma$; error budget $\varepsilon$ & $16$; $4$ tokens of the previous view \\
 & maximum magnification $k_{\max}$ (linear) & $4$ \\
 & first-stage looseness $\lambda$; maximum depth $D$ & $2$; $2$ stages \\
 & minimum crop side $d_{\min}$ & $192$\,px \\
\midrule
Verifier & window & side $d_0(W,H)$ centred on the candidate, shifted inside the screenshot, $4\times$ area \\
 & mark & red circle of radius $r=\max(10,\lfloor0.025L\rfloor)$, stroke $\max(2,\lfloor r/4\rfloor)$ \\
\midrule
Chain arbitration & portfolio $\mathcal{C}$ (priority order) & $\{c_{\mathrm{TNZ}},c_{4\times}\}$ \\
 & fixed chain $c_{4\times}$ & whole screen at $b=1$ $\to$ $1288{\times}728$-px window around its answer, $4\times$ area \\
 & grouping radius $\delta_{\text{px}}$; representatives $K$ & $14$\,px; $3$ \\
\midrule
Re-grounding & threshold $\theta_r$ & 0.99 \\
 & schedule & whole screen at $b=1$ with the rejected point crossed out $\to$ native crop of side $d_0$ $\to$ crop of side $d_0/2$ at $9\times$ area \\
\midrule
Abstention & rule; threshold $\theta_e$ & abstain iff $\tfrac12(v(p)+e(S,I))<\theta_e$; $0.03$; never on a CSG-composed answer; OSWorld-G only \\
\midrule
CSG & operators & \textsc{ground}, \textsc{cell}$(\mathit{col},\mathit{row})$ \\
 & header refinement & whole screen at $b=1$ $\to$ crop of side $d_0/2$ at $9\times$ area \\
\bottomrule
\end{tabular}
\caption{\textbf{Constants of AdaGround} (Section~\ref{sec:config}). A single set serves both backbones and all benchmarks, and only $\pi$ is read off the encoder. $d_0(W,H)$ is the window side of Section~\ref{sec:verify}, $L$ is the longer side of the rendered view, and $n\times$ area means $\sqrt{n}$ times the linear size.}
\label{tab:constants}
\end{table*}

\paragraph{Baselines and the fixed chain $c_{4\times}$.} The \emph{single pass} grounds on the whole screenshot at $b=1$. The \emph{fixed $4\times$ crop}, i.e.\ the chain $c_{4\times}$, re-grounds on a $1288{\times}728$-pixel window centred on the single-pass answer (clamped to the screen size, shifted inside the screenshot) at four times its area. Both baselines come from the same runs as AdaGround. The verifier, re-grounding and CSG windows are squares of side $d_0(W,H)$ or $d_0/2$, where $d_0$ is $512$ pixels on a $1920{\times}1080$ screen and $1{,}018$ on a $3840{\times}2160$ one.

\paragraph{Development and held-out evaluation.} All components and constants, including the $1288{\times}728$ window, were developed on OSWorld-G, ScreenSpot-Pro and ScreenSpot-v2, mostly with Qwen3-VL-8B, by comparing alternatives on these benchmarks. The abstention rule and $\theta_e$ were chosen on the full OSWorld-G. No part of the three was held out, so their numbers include the effect of these choices. Varying $\tau\in\{2,3,4\}$ (Qwen3-VL-8B) or $\theta_r\in\{0.9,0.99\}$ changes accuracy by at most $0.6$ points (Appendix~\ref{sec:app_variants}). The configuration was neither developed nor tuned on UI-Vision~\citep{uivision2025}, which is evaluated only with the final constants (Table~\ref{tab:uivision}) and the fixed-geometry control (Appendix~\ref{sec:app_controls}). An earlier version of the method had been evaluated on it.

\paragraph{Verifier score along the pipeline.} The decisions of Section~\ref{sec:verify} give the following property.
\begin{property}[Monotone verification]\label{prop:monotone}
Treat $v$ as a deterministic function of $(S,I,p)$. Up to the CSG step, the verifier score of the current answer never decreases wherever it is evaluated (a property of the score, not of accuracy).
\end{property}
\noindent\emph{Argument.} TNZ accepts only a non-decreasing $v$ (Eq.~\eqref{eq:accept}), and re-grounding accepts only a strictly higher one. Since $|\mathcal{C}|\le K$, arbitration verifies every group and returns the arg max over a set that contains the $c_{\mathrm{TNZ}}$ answer. A single consensus group keeps its $c_{\mathrm{TNZ}}$ member.\hfill$\square$

\begin{algorithm*}[t]
\caption{AdaGround}
\label{alg:adaground}
\small
\renewcommand{\algorithmiccomment}[1]{\hfill$\triangleright$~#1}%
\begin{algorithmic}[1]
\REQUIRE screenshot $S$ ($H\times W$), instruction $I$, VLM $M$, token pitch $\pi$, constants of Section~\ref{sec:config}, portfolio $\mathcal{C}=\{c_{\mathrm{TNZ}},c_{4\times}\}$ in priority order
\ENSURE click point $p$ in screenshot pixels, or \textsc{abstain}
\STATE $\rho\leftarrow\textsc{Rewrite}_M(I)$ \COMMENT{CSG: text-only program, $\rho\in\{\textsc{ground},\textsc{cell}(\mathit{col},\mathit{row})\}$}
\STATE $p^{(c)}\leftarrow c(S,I)$ for every $c\in\mathcal{C}$, with $c_{\mathrm{TNZ}}=\textsc{TNZ}(S,I)$ below \COMMENT{VLR: chain portfolio}
\STATE $\mathcal{G}\leftarrow$ complete-linkage groups of $\{p^{(c)}\}$ at radius $\delta_{\text{px}}$; $r(G)\leftarrow$ highest-priority member of $G$
\STATE \textbf{if} $|\mathcal{G}|\ge2$ \textbf{then} $p\leftarrow\arg\max_{G\in\mathcal{G}_K}v(r(G))$ over the $K$ largest groups \textbf{else} $p\leftarrow r(G)$ \COMMENT{Eq.~\eqref{eq:arbitrate}}
\IF{$v(p)<\theta_r$}
\STATE $p''\leftarrow\textsc{Reground}(S,I,p)$: whole screen, $p$ crossed out $\to$ native $d_0$ crop $\to$ $d_0/2$ crop at $3\times$ linear \COMMENT{VLR: negative feedback}
\STATE \textbf{if} $v(p'')>v(p)$ \textbf{then} $p\leftarrow p''$
\ENDIF
\IF{$\rho=\textsc{cell}(\mathit{col},\mathit{row})$}
\STATE ground both header descriptions (whole screen, then $d_0/2$ crop at $3\times$ linear) $\rightarrow(x_{\mathrm{col}},y_{\mathrm{col}}),(x_{\mathrm{row}},y_{\mathrm{row}})$ \COMMENT{CSG}
\STATE \textbf{if} $x_{\mathrm{col}}>x_{\mathrm{row}}$ \textbf{and} $y_{\mathrm{col}}<y_{\mathrm{row}}$ \textbf{then return} $(x_{\mathrm{col}},y_{\mathrm{row}})$ \COMMENT{Eq.~\eqref{eq:cell}}
\ENDIF
\STATE \textbf{if} refusal admissible \textbf{and} $\frac12(v(p)+e(S,I))<\theta_e$ \textbf{then return} \textsc{abstain} \COMMENT{VLR: Eq.~\eqref{eq:abstain}}
\RETURN $p$
\end{algorithmic}
\vspace{1pt}\hrule\vspace{2pt}
\renewcommand{\algorithmicrequire}{\textbf{Procedure}}
\begin{algorithmic}[1]
\REQUIRE $\textsc{TNZ}(S,I)$ \COMMENT{Token-Normalized Adaptive Zooming with gated acceptance}
\STATE $b_1\leftarrow\min(1,\max(b_{\min},A_0/(HW)))$;\; $p\leftarrow M(\mathcal{R}(S,b_1),I)$;\; $\hat s\leftarrow$ shorter side of $\textsc{Box}_M(\mathcal{R}(S,b_1),I)$ in original pixels;\; $k_{\text{prev}}\leftarrow\sqrt{b_1}$
\FOR{$t=1$ \TO $D$}
\STATE $\hat\sigma\leftarrow\hat s\,k_{\text{prev}}/\pi$
\STATE \textbf{if} no coarse box \textbf{or} $\hat\sigma\ge\lambda_t\tau$ \textbf{or} ($t\ge2$ \textbf{and} $k_{\text{prev}}\ge k_{\max}$) \textbf{then break} \COMMENT{adaptive depth}
\STATE $k\leftarrow$ Eq.~\eqref{eq:k};\; $d\leftarrow$ Eq.~\eqref{eq:adaptive_d};\; $C\leftarrow$ box of side $d$ centred on $p$, shifted inside $S$, top-left $(x_0,y_0)$
\STATE $p'\leftarrow(x_0,y_0)+M(\mathcal{R}(C,k^2),I)/k$ \COMMENT{Eq.~\eqref{eq:map_back}}
\STATE \textbf{if} $t<D$ \textbf{then} $\hat s\leftarrow$ shorter side of $\textsc{Box}_M(\mathcal{R}(C,k^2),I)$ in original pixels (kept if no box)
\STATE \textbf{if} $v(p')\ge v(p)$ \textbf{then} $p\leftarrow p'$;\; $k_{\text{prev}}\leftarrow k$ \COMMENT{gated acceptance, Eq.~\eqref{eq:accept}}
\ENDFOR
\RETURN $p$
\end{algorithmic}
\end{algorithm*}

\subsection{Implementation Details}
\label{sec:implementation}

\begin{itemize}[leftmargin=*,nosep]
    \item \textbf{Backbones and inference engine:} Qwen3-VL-8B-Instruct and Qwen3-VL-32B-Instruct (FP8 weights) on vLLM~$0.11.2$, greedy decoding (temperature $= 0$), \texttt{enforce\_eager=True}, $4{\times}$NVIDIA H20 GPUs.
    \item \textbf{Prompt template:} Qwen3-VL's chat template with the grounding system message of Appendix~\ref{sec:prompts_verbatim}, which states the size of the rendered view. The model writes points on the $[0,1000]$ grid of that view. This is its native convention, and the prompt does not restate it. A grounding pass generates at most $32$ tokens.
    \item \textbf{Rendering:} every view is resized to a $32$-pixel grid with $3{,}136$ to $8{,}847{,}360$ pixels. A view at area factor $b>1$ is first up-sampled by $\sqrt b$ in each dimension. A coarse view with $b<1$ is capped at $b$ times the pixel count of the full-resolution view.
    \item \textbf{Coarse pass:} the budget $b_1$ of Section~\ref{sec:zoom} renders a $1920{\times}1080$ screenshot at native resolution and a $3840{\times}2160$ one at about half its linear size. The point query and the box query are two separate calls on the same coarse view. If no box is returned, no zoom is issued.
    \item \textbf{TNZ geometry:} Eqs.~\eqref{eq:k} and~\eqref{eq:adaptive_d} with the constants of Table~\ref{tab:constants}. The $d\times d$ crop is centred on the current answer, shifted inside the screenshot and rendered at $k^2$ times its area. The box is re-queried on the zoomed view after every stage except the last, and $\hat s$ is updated whether or not the stage is accepted. A zoomed answer $p'$ replaces $p$ iff $v(p')\ge v(p)$, in which case $k_{\text{prev}}\leftarrow k$.
    \item \textbf{Verifier-in-the-loop refinement:} $v(p)$ is read from the red circle of Table~\ref{tab:constants}, drawn at the candidate on its own window of side $d_0(W,H)$ and rendered at four times the area. This window is independent of the crop of the stage being verified. The score is $v(p)=P(\text{Yes})/(P(\text{Yes})+P(\text{No}))$, computed from the first-token log-probabilities with the case variants of each word summed over the $20$ most likely first tokens. A word outside them contributes zero, and $v=0.5$ when neither word is among them. Re-grounding fires when $v(p)<\theta_r$ and keeps its answer only if that answer verifies strictly higher. Arbitration groups the chains' answers at $\delta_{\text{px}}=14$ pixels and verifies up to $K=3$ group representatives. A tie goes to $c_{\mathrm{TNZ}}$. The existence query is one whole-screen Yes/No pass at $b=1$ whose normalized \textsc{Yes} probability is $e(S,I)$. OSWorld-G is the only benchmark here that admits a refusal. On it, the pipeline abstains iff $\tfrac12(v(p)+e(S,I))<\theta_e=0.03$ (Eq.~\eqref{eq:abstain}), where $v(p)$ is the score of the answer after re-grounding. It never abstains on a CSG-composed answer.
\end{itemize}

\subsection{Prompt Template and Output Parsing}
\label{sec:prompt_template}

\noindent\textbf{Template structure.} Every grounding pass uses the grounding system prompt of Appendix~\ref{sec:prompts_verbatim}, one image and the benchmark instruction $I$ verbatim. The image is the whole screenshot or a crop, at the area factor of its stage, and no chain-of-thought or reasoning suffix is added. The box, verification, existence and program queries use the system message ``You are a helpful assistant.'' and the strings of Appendix~\ref{sec:prompts_verbatim}.

\noindent\textbf{Output format and parsing.} The point query returns a pair $(\alpha,\beta)$ on the grid $[0, 1{,}000]^{2}$ of the rendered view. The parser takes the first pair written as $(x,y)$, provided every such pair has integer values. It maps this pair to the screenshot as $(\lfloor x_0+\alpha\,w_C/1000\rfloor,\ \lfloor y_0+\beta\,h_C/1000\rfloor)$, where $(x_0,y_0)$ and $w_C\times h_C$ are the crop's actual corner and size in original pixels (the whole screenshot for a whole-screen view). This is Eq.~\eqref{eq:map_back} without its dependence on the rendered size, so neither the rounding to the $32$-pixel grid nor the pixel cap enters the mapping. Every crop and window is centred on the current answer and shifted inside the screenshot, hence the actual corner rather than $p-d/2$. A pass without such a pair yields no point, and its stage keeps the answer it had. Accordingly, TNZ does not zoom when the coarse pass fails, a zoom stage keeps its starting answer, re-grounding keeps the answer of its previous step, and a header grounding keeps its whole-screen point. The box query returns \texttt{"bbox\_2d": [x$_1$, y$_1$, x$_2$, y$_2$]} on the same grid. Its shorter side, rescaled to original pixels, is $\hat s$.

\noindent\textbf{Determinism and reproducibility.} Outputs are not bit-exact across runs, because the continuous batching of vLLM~\citep{vllm2023} can change the batch in which a request is computed. Every configuration can be reproduced from public model weights, public benchmark inputs and the settings above.

\subsection{Verbatim Prompts}
\label{sec:prompts_verbatim}
Every decision in AdaGround is a query to the same backbone $M$. The re-grounding threshold $\theta_r$ (Table~\ref{tab:constants}) and the existence decision are defined on the first-token probabilities that these exact strings produce, so we give the strings in full. In them, \emph{[instruction]} stands for the benchmark instruction, inserted verbatim (in the rewrite query its double quotes are replaced by single quotes). The symbol $\hookleftarrow$ marks a line break, and quotation marks inside the prompts are typeset.

\noindent\textbf{Grounding (system message).} \emph{[H]} and \emph{[W]} are the height and width of the rendered view.
\begin{quote}\small
``You are an expert UI element locator. Given a GUI image and a user's element description, provide the coordinates of the specified element as a single (x,y) point. The image resolution is height \emph{[H]} and width \emph{[W]}. For elements with area, return the center point.\,$\hookleftarrow\hookleftarrow$ Output the coordinate pair exactly:\,$\hookleftarrow$ (x,y)''
\end{quote}

\noindent\textbf{Size estimation (box query).} The system message is ``You are a helpful assistant.'' and the user text is:
\begin{quote}\small
``Locate the UI element that this instruction refers to: "\emph{[instruction]}".\,$\hookleftarrow$ Output its bounding box as JSON \texttt{[\{"bbox\_2d": [x1, y1, x2, y2]\}]} with coordinates on the 0-1000 grid of the image.''
\end{quote}

\noindent\textbf{Verification $v(p)$.}
\begin{quote}\small
``The red circle marks a candidate click position in this region of a screenshot. Instruction: "\emph{[instruction]}". Is the red circle on the UI element that this instruction refers to? Answer Yes or No.''
\end{quote}

\noindent\textbf{Existence $e(S,I)$.}
\begin{quote}\small
``Instruction: "\emph{[instruction]}".\,$\hookleftarrow$ Is there a UI element on this screen that this instruction refers to, so that the instruction can be carried out by clicking it? Answer Yes or No.''
\end{quote}

\noindent\textbf{Negative-feedback re-grounding.} This query uses the grounding prompt above on the whole-screen view, with a red cross drawn at the rejected location and this note appended to the instruction:
\begin{quote}\small
``Note: a previous attempt clicked the location marked with the red X, and that was judged NOT to be the target. Find the correct element (it may be close to or far from the X).''
\end{quote}
On a view of longer side $L$, the cross has arms reaching $\max(8,\lfloor0.012\,L\rfloor)$ pixels from the centre along each axis. The cross and the note are used in the first, whole-screen pass only. The two crops that follow use the plain instruction.

\noindent\textbf{CSG: instruction-to-program rewrite.} A text-only query, with no image:
\begin{quote}\small
``Convert a GUI click instruction into a JSON program for a grounding engine. Operations:\,$\hookleftarrow$
\texttt{1. \{"op": "ground"\}} - the instruction names or describes a UI element (button, icon, menu item, tab, field, link, list item, text, image, etc.) and clicking anywhere on that element is fine. Use this for almost all instructions.\,$\hookleftarrow$
\texttt{2. \{"op": "cell", "col": "<column letters>", "row": <row number>\}} - the instruction asks for one cell inside the data grid of a spreadsheet, given by its column and row, e.g.\ ``cell C12'' or ``the 3rd column and 12th row'' -> \texttt{\{"op": "cell", "col": "C", "row": 12\}}. Do NOT use it for a row header (row number button), a column header (column letter button), a cell corner or fill handle, a cell described only by its content, or a grid that is not a spreadsheet (color palette, shape gallery, icon table, calendar) - use ground for those.\,$\hookleftarrow$
\texttt{3. \{"op": "along", "element": "<short description of the long element>", "fx": <0..1>, "fy": <0..1>\}} - the instruction asks for a specific relative position on a slider, progress bar, scroll bar, divider line or another long thin element; fx/fy are the relative position inside that element's box, e.g.\ ``the left end of the volume slider'' -> fx 0.02, fy 0.5; ``30\% along the seek bar'' -> fx 0.3, fy 0.5; ``the bottom part of the vertical scroll bar'' -> fx 0.5, fy 0.9; ``the middle of the divider'' -> fx 0.5, fy 0.5.\,$\hookleftarrow$
\texttt{4. \{"op": "text\_edge", "anchor": "<exact visible text>", "side": "before" or "after", "context": "<where that text is>"\}} - the instruction asks to place the text cursor immediately before or after a specific visible piece of text, e.g.\ ``put the cursor right after the word 'Total:' in the table header'' -> \texttt{\{"op": "text\_edge", "anchor": "Total:", "side": "after", "context": "table header"\}}.\,$\hookleftarrow$
Instruction: "\emph{[instruction]}"\,$\hookleftarrow$ Output only the JSON object.''
\end{quote}
Programs of the last two types are not executed and fall back to \textsc{ground}. The \textsc{cell} operator grounds two header instructions, ``The column header labeled "\emph{[col]}" at the top of the spreadsheet grid'' and ``The row header labeled "\emph{[row]}" at the left of the spreadsheet grid''. Each is grounded first on the whole screen at $b=1$ and then on a crop of side $d_0/2$ around that point at $9\times$ area. The answer $(x_{\text{col}},y_{\text{row}})$ is composed iff $x_{\text{col}}>x_{\text{row}}$ and $y_{\text{col}}<y_{\text{row}}$. Otherwise the VLR answer is kept.

\noindent\textbf{Chain arbitration.} The chains' answers are grouped by complete linkage at $\delta_{\text{px}}$, so a point joins a group only if it lies within $\delta_{\text{px}}$ of every member. Each group is represented by its member from the highest-priority chain. With a single group, its representative is returned. Otherwise the representatives of up to $K$ largest groups are verified independently with the verification prompt above, each on its own window, and the highest $v$ wins. No prompt shows two candidates at once.

\subsection{Compute and Cost}\label{sec:app_compute}
All experiments ran on one node with $4\times$ NVIDIA H20 (96\,GB). AdaGround and the baselines ran under vLLM~$0.11.2$ with batched offline inference (at most $16$ and $8$ concurrent sequences), BF16 weights for Qwen3-VL-8B and FP8 weights for Qwen3-VL-32B. In Table~\ref{tab:cost}, re-grounding is charged its three passes and their verification only where it fires, and the verification that triggers it on every sample. Latency is measured with sequential calls on one GPU at batch size $1$, on $150$ random samples per benchmark, with vLLM's prefix caching enabled (as in all experiments) and each stage's warm-up excluded. The text-only CSG rewrite is sent as one batched request and charged at its mean share per sample. AdaGround takes $3.4$ to $4.2$ times the latency of the fixed $4\times$ crop. On both backbones it reads $3.5$ to $3.9$ times the visual tokens of that crop ($1.7$ to $2.0$ times those of $c_{\mathrm{TNZ}}$ alone). On Qwen3-VL-32B it makes $12.2$ to $13.7$ calls per sample, $6.2$ to $6.6$ of them in sequence.
\begin{table}[t]
\centering
\scriptsize
\setlength{\tabcolsep}{2.5pt}
\resizebox{\columnwidth}{!}{\begin{tabular}{@{}lcccccc@{}}
\toprule
\textbf{Qwen3-VL-8B} & Acc. & Tok.\,(k) & Calls & Depth & Lat.\,(s) & p95 \\
\midrule
\multicolumn{7}{@{}l}{\textit{OSWorld-G}} \\
Single pass & 64.7 & 1.7 & 1.0 & 1.0 & 0.9 & 1.3 \\
Fixed $4\times$ crop & 70.7 & 5.3 & 2.0 & 2.0 & 3.1 & 3.8 \\
$c_{\mathrm{TNZ}}$ & 72.7 & 10.4 & 6.9 & 4.0 & 7.7 & 10.5 \\
\quad + arbitration & 73.6 & 16.2 & 9.3 & 4.2 & 11.1 & 14.6 \\
\quad + re-grounding & 73.2 & 17.7 & 10.9 & 5.8 & 12.0 & 16.5 \\
\quad + CSG & 73.6 & 17.9 & 12.0 & 5.8 & 12.7 & 17.2 \\
\quad + abstention & 76.1 & 19.5 & 13.0 & 5.8 & 13.3 & 18.3 \\
\multicolumn{7}{@{}l}{\textit{ScreenSpot-Pro}} \\
Single pass & 51.3 & 5.4 & 1.0 & 1.0 & 4.7 & 7.6 \\
Fixed $4\times$ crop & 63.5 & 9.1 & 2.0 & 2.0 & 7.3 & 10.3 \\
$c_{\mathrm{TNZ}}$ & 59.6 & 15.7 & 6.9 & 4.0 & 11.8 & 18.8 \\
\quad + arbitration & 65.6 & 27.3 & 9.8 & 4.5 & 21.0 & 33.1 \\
\quad + re-grounding & 66.2 & 31.9 & 11.4 & 6.2 & 24.5 & 45.5 \\
\quad + CSG & 66.2 & 31.9 & 12.5 & 6.2 & 24.9 & 45.8 \\
\multicolumn{7}{@{}l}{\textit{ScreenSpot-v2}} \\
Single pass & 92.1 & 2.7 & 1.0 & 1.0 & 1.8 & 2.6 \\
Fixed $4\times$ crop & 91.9 & 6.0 & 2.0 & 2.0 & 4.2 & 5.2 \\
$c_{\mathrm{TNZ}}$ & 92.5 & 12.2 & 6.9 & 4.0 & 8.9 & 11.3 \\
\quad + arbitration & 92.5 & 18.8 & 9.3 & 4.3 & 13.3 & 17.2 \\
\quad + re-grounding & 91.9 & 21.0 & 10.8 & 5.8 & 14.6 & 21.1 \\
\quad + CSG & 91.9 & 21.0 & 11.8 & 5.8 & 14.9 & 21.4 \\
\bottomrule
\end{tabular}}
\caption{\textbf{Cost per sample} (Qwen3-VL-8B, full benchmarks). Columns: accuracy (\%), visual tokens (thousands), model calls, and calls in sequence (depth, with independent calls in parallel). Latency: mean and 95th-percentile seconds per sample on $150$ random samples per benchmark, with sequential calls (Appendix~\ref{sec:app_compute}). The last row of each block is AdaGround.}
\label{tab:cost}
\end{table}

\section{Ablations and Component Analyses}\label{sec:app_ablation}
All results use the full benchmarks and, except for the quantity varied, the configuration of Section~\ref{sec:config}. Intervals are computed as in Appendix~\ref{sec:statistics}.

\subsection{Components}\label{sec:extended_ablation}
This subsection tests the steps of Table~\ref{tab:ladder}.

\paragraph{TNZ.} $c_{\mathrm{TNZ}}$ improves the single pass by $4.8$ to $17.1$ points on OSWorld-G and ScreenSpot-Pro and stays within a point of it on ScreenSpot-v2 ($+0.4$ and $+0.6$, both intervals containing zero). Compared with the fixed $4\times$ crop, it is significantly lower on Qwen3-VL-8B ScreenSpot-Pro ($-3.9$ $[-5.8,-2.0]$), and the other five differences ($-1.5$ to $+2.0$ points) are not significant. On held-out UI-Vision, $c_{\mathrm{TNZ}}$ is above the fixed crop by $+2.6$ $[+1.5,+3.6]$ and $+2.5$ $[+1.2,+3.8]$ points. There, the full AdaGround's lead over the fixed crop is largest on spatial instructions, which name the target by its relation to other elements ($+4.9$ and $+7.9$).

\paragraph{Chain arbitration} pays off where the chains' answers fall into two groups. This happens on $716$ ($45\%$) and $545$ ($34\%$) of the $1{,}581$ ScreenSpot-Pro samples (Qwen3-VL-8B and 32B), but only on $115$ and $107$ of the $564$ OSWorld-G samples. On ScreenSpot-Pro, arbitration raises $c_{\mathrm{TNZ}}$ by $+5.9$ $[+4.6,+7.3]$ and $+3.2$ $[+2.2,+4.4]$ points, recovering $38.8\%$ and $35.9\%$ of the headroom between the better chain and the oracle (Table~\ref{tab:arbitration}). Most of AdaGround's gain over the fixed crop there is thus already present after arbitration ($2.1$ and $1.8$ of $2.7$ and $2.0$ points, Table~\ref{tab:paired}). On OSWorld-G it adds $+0.9$ $[-0.3,+2.2]$ and $+0.2$ $[-0.4,+0.8]$ points over $c_{\mathrm{TNZ}}$, and on ScreenSpot-v2 $0.0$ and $+0.6$ $[+0.2,+1.2]$. Many verifier decisions are between scores near $1$. Both scores are at least $0.999$ in $45.2$--$76.9\%$ of TNZ acceptance decisions, and the two highest scores are in $10.3$--$47.2\%$ of arbitrations.

\paragraph{Re-grounding} helps on ScreenSpot-Pro but costs $0.6$ points on ScreenSpot-v2, where more than $92\%$ of answers are already right and it replaces more correct answers than it repairs.

\paragraph{CSG} acts only on OSWorld-G, where it adds $2$ and $5$ samples net (not significant). Qwen3-VL-8B and 32B rewrite $20$ and $23$ instructions into cell programs and compose a cell on $19$ of each. On these programs the chain up to re-grounding solves $13$ of each, the single pass $15$ and $17$, and AdaGround $15$ and $18$. Returning the single-pass answer instead of composing (exempt from abstention, like a composed answer) does as well. It changes AdaGround by $0.0$ $[-0.5,+0.5]$ and $-0.2$ $[-1.1,+0.5]$ points, and on the composed cells it is right on $15$ and $16$, against $14$ and $15$ for the composed point. CSG's gain thus comes from replacing the zoomed answer, not from composing the headers.

\paragraph{Abstention} adds $2.5$ and $1.4$ points on OSWorld-G (Table~\ref{tab:ladder}) and accounts for about half or more of AdaGround's gain there over the fixed crop ($14$ of $30$ and $8$ of $12$ samples net). Without it, the lead over the crop is significant on Qwen3-VL-8B only (Table~\ref{tab:paired}, superscript~$a$).

\subsection{Abstention}\label{sec:mech_validation}
\begin{table}[t]
\centering
\scriptsize
\setlength{\tabcolsep}{3pt}
\renewcommand{\arraystretch}{0.9}
\begin{tabular}{@{}lllccc@{}}
\toprule
\textbf{Backbone} & \textbf{Rule} & $\theta_e$ & \textbf{Overall} & \makecell{\textbf{Present}\\\textbf{targets}} & \makecell{\textbf{Abstentions}\\(correct / false)} \\
\midrule
Qwen3-VL-8B & no gate & -- & 73.6 & 81.4 & 0 / 0 \\
 & existence & 0.01 & 74.6 & 77.8 & 24 / 31 \\
\rowcolor{table_color!20}  & joint & 0.03 & 76.1 & 81.2 & 15 / 4 \\
\midrule
Qwen3-VL-32B$^{\ast}$ & no gate & -- & 77.8 & 86.1 & 0 / 0 \\
 & existence & 0.01 & 81.9 & 83.9 & 34 / 19 \\
\rowcolor{table_color!20}  & joint & 0.03 & 79.3 & 86.1 & 8 / 2 \\
\bottomrule
\end{tabular}
\caption{\textbf{Abstention rules on OSWorld-G}. \emph{Existence}: abstain iff $e(S,I)<\theta_e$. \emph{Joint}: abstain iff $\tfrac12(v(p)+e(S,I))<\theta_e$ (Eq.~\eqref{eq:abstain}). Abstentions: correct (absent target) / false (present target). Present targets: accuracy where the target is present. The existence rule uses $\theta_e=0.01$, the most conservative value tried. Shaded: the configuration used elsewhere. $^{\ast}$FP8 weights.}
\label{tab:gate_ablation}
\end{table}

The existence query alone withdraws many answers whose target is present (Table~\ref{tab:gate_ablation}). The joint rule of Eq.~\eqref{eq:abstain} abstains only when the verifier also fails to confirm the answer. Its overall accuracy stays within $75.9$--$76.2\%$ and $79.1$--$79.6\%$ for $\theta_e$ from $0.01$ to $0.2$. At $\theta_e=0.03$ it recalls $27.8\%$ $[17.6,40.9]$ and $14.8\%$ $[7.7,26.6]$ of the absent targets at false-refusal rates of $0.8\%$ $[0.3,2.0]$ and $0.4\%$ $[0.1,1.4]$ (Wilson intervals). On Qwen3-VL-32B the existence rule at $\theta_e=0.01$ reaches a higher overall accuracy, but with $19$ false abstentions.

The rule also works without AdaGround. Applied to the answers of the fixed $4\times$ crop and of MVP on OSWorld-G (Table~\ref{tab:paired}, superscript~$b$), it raises them to $73.8\%$ ($+3.0$ $[+1.3,+5.0]$) and $73.0\%$ ($+3.2$ $[+1.5,+5.0]$) on Qwen3-VL-8B. On 32B it raises the fixed crop to $78.5\%$ ($+1.4$ $[+0.2,+2.8]$).

\subsection{Quality of the Size Estimate}\label{sec:shat_quality}
Both backbones return a box, and thus $\hat s$, on every sample. On ScreenSpot-Pro the estimate is coarse. Its median ratio to the true side is $1.2$ (Qwen3-VL-8B) and $1.4$ (32B), and its error is below the true side on $82.4\%$ and $84.1\%$ of samples. Both backbones therefore under-magnify on average (mean first-stage magnification $3.1$ and $2.7$, against $3.4$ with the true $s$). On ScreenSpot-v2 the estimate is closer (median ratio $0.8$ and $0.9$) and both over-magnify slightly ($1.8$ and $1.7$ against $1.5$). TNZ is robust to this error. With the true $s$, the arbitrated accuracy changes by $-0.6$ to $-0.2$ points, and every interval contains zero (Qwen3-VL-8B). The estimate is not corrected, and a target outside an over-tight view can only be rejected by verification, not recovered.

\subsection{Design Choices}\label{sec:app_variants}
Removing one component of AdaGround at a time gives three significant differences, all of them losses. They come from abstention on OSWorld-G with both backbones and from re-grounding on Qwen3-VL-8B ScreenSpot-Pro. A looser re-grounding threshold ($\theta_r=0.9$) changes accuracy by at most one sample net in every cell. Accuracy is not sensitive to the target span and changes by at most $0.6$ points at $\tau=2$ and $\tau=4$. Within TNZ, the first zoom carries most of the gain over the coarse point. The negative feedback in re-grounding (the crossed-out location and the exclusion note) changes which answers re-grounding returns, but overall accuracy differs significantly on no benchmark.

\subsection{Chain Arbitration}\label{sec:app_arbitration}
\begin{table*}[t]
\centering
\footnotesize
\setlength{\tabcolsep}{4pt}
\renewcommand{\arraystretch}{1.0}
\begin{tabular*}{\textwidth}{@{}l@{\extracolsep{\fill}}cccccccc@{}}
\toprule
 & \multicolumn{5}{c}{\textbf{Accuracy (\%)}} & \multicolumn{2}{c}{\textbf{Arbitration minus (points)}} & \textbf{Verifier} \\
\cmidrule(lr){2-6}\cmidrule(lr){7-8}\cmidrule(l){9-9}
\textbf{Benchmark} & $c_{\mathrm{TNZ}}$ & $c_{4\times}$ & random & arbitration & oracle & better chain & random & pick \\
\midrule
\multicolumn{9}{@{}l}{\textit{Qwen3-VL-8B}} \\
OSWorld-G & \underline{72.7} & 70.7 & 71.7 & 73.6 & 76.1 & $+0.9$ $[-0.3,+2.2]$ 8/3 & $+1.9$ $[+0.6,+3.1]$ & 27/35 \\
ScreenSpot-Pro & 59.6 & \underline{63.5} & 61.6 & 65.6 & 68.9 & $+2.1$ $[+0.8,+3.3]$ 67/34 & $+4.0$ $[+3.1,+4.9]$ & 177/227 \\
ScreenSpot-v2 & \underline{92.5} & 91.9 & 92.2 & 92.5 & 94.0 & $0.0$ $[-0.6,+0.6]$ 8/8 & $+0.3$ $[-0.2,+0.8]$ & 27/46 \\
\midrule
\multicolumn{9}{@{}l}{\textit{Qwen3-VL-32B$^{\ast}$}} \\
OSWorld-G & 76.6 & \underline{77.1} & 76.9 & 76.8 & 79.8 & $-0.4$ $[-2.1,+1.5]$ 13/15 & $-0.1$ $[-1.0,+0.9]$ & 11/24 \\
ScreenSpot-Pro & 72.4 & \underline{73.9} & 73.1 & 75.6 & 78.8 & $+1.8$ $[+0.4,+3.0]$ 65/37 & $+2.5$ $[+1.7,+3.4]$ & 126/174 \\
ScreenSpot-v2 & 95.4 & \underline{95.4} & 95.4 & 96.0 & 96.6 & $+0.6$ $[-0.1,+1.2]$ 13/6 & $+0.6$ $[+0.2,+1.0]$ & 23/31 \\
\bottomrule
\end{tabular*}
\caption{\textbf{Chain arbitration} before re-grounding, CSG and abstention. Random: either chain with probability $\frac12$ (expected). Oracle: either chain is right. Underlined: the better chain, chosen after the fact. Differences: $95\%$ CI and, against the better chain, samples won/lost. Pick: verifier picks the right chain / samples with exactly one chain right (verifier called). $^{\ast}$FP8 weights.}
\label{tab:arbitration}
\end{table*}

Arbitration (Eq.~\eqref{eq:arbitrate}) is significantly above the better chain on ScreenSpot-Pro, and significantly above a random choice in four of the six cells (Table~\ref{tab:arbitration}). The gain comes from the samples on which exactly one chain is right. On Qwen3-VL-8B ScreenSpot-Pro, both chains are right on $858$ samples, only $c_{4\times}$ on $146$, only $c_{\mathrm{TNZ}}$ on $85$ and neither on $492$. The verifier is called on the $716$ samples whose two clicks lie more than $\delta_{\text{px}}$ apart ($545$ on Qwen3-VL-32B). These include $227$ of the $231$ samples with exactly one chain right. The other $4$ keep $c_{\mathrm{TNZ}}$'s answer. On these $227$, always taking $c_{4\times}$ would be right on $144$ ($63.4\%$), while the verifier is right on $177$ ($78.0\%$). On Qwen3-VL-32B, the counts are $99$ of $174$ ($56.9\%$) against $126$ ($72.4\%$).

\paragraph{Ties and score precision.} On Qwen3-VL-8B, $304$ of the $1{,}066$ arbitrations ($28.5\%$) are ties at four decimals. Most of them have both scores at $1.0000$, i.e.\ the verifier confirms both clicks. On 32B the tie rate is $9.8\%$, and fewer than half of these ties are at $1.0000$. Where exactly one chain is right on a tie, the verifier picks it on $28$ of $56$. The net gain over $c_{4\times}$ comes from decisions with a margin above $10^{-4}$ ($+61$ samples, against $-6$ at smaller margins). The score uses the $20$ most likely first tokens, which contain Yes and No in all $8{,}762$ verifications. Scoring with the $1{,}000$ most likely tokens changes no answer's correctness.

\subsection{Fixed-Geometry Controls}\label{sec:app_controls}
\begin{table*}[t]
\centering
\footnotesize
\setlength{\tabcolsep}{4pt}
\resizebox{\ifdim\width>\textwidth\textwidth\else\width\fi}{!}{\begin{tabular}{@{}llcccc@{}}
\toprule
& \textbf{Second chain} & \textbf{Chain} & \textbf{+ arbitration} & \textbf{AdaGround variant} & $\Delta$ \textbf{vs.\ AdaGround} \\
\midrule
\multicolumn{6}{@{}l}{\textit{Qwen3-VL-8B}} \\
OSWorld-G & $c_{\mathrm{TNZ}}$ (AdaGround) & 72.7 & 73.6 & 76.1 & -- \\
 & fixed $540{\times}540$ crop, $2.78\times$ & 71.1 ($-1.6$) & 72.7 ($-0.9$ $[-2.9,+1.2]$) & 75.7 & $-0.4$ $[-2.3,+1.5]$ \\
\addlinespace[3pt]
ScreenSpot-Pro & $c_{\mathrm{TNZ}}$ (AdaGround) & 59.6 & 65.6 & 66.2 & -- \\
 & fixed $540{\times}540$ crop, $2.78\times$ & 62.4 ($+2.7$) & 65.7 ($+0.1$ $[-1.1,+1.3]$) & 66.1 & $-0.1$ $[-1.3,+1.1]$ \\
\addlinespace[3pt]
ScreenSpot-v2 & $c_{\mathrm{TNZ}}$ (AdaGround) & 92.5 & 92.5 & 91.9 & -- \\
 & fixed $540{\times}540$ crop, $2.78\times$ & 89.3 ($-3.2$) & 90.8 ($-1.7$ $[-2.8,-0.7]$) & 91.0 & $-0.9$ $[-1.9,0.0]$ \\
\addlinespace[3pt]
UI-Vision$^{h}$ & $c_{\mathrm{TNZ}}$ (AdaGround) & 30.4 & 31.3 & 31.7 & -- \\
 & fixed $540{\times}540$ crop, $2.78\times$ & 29.1 ($-1.2$) & 30.2 ($-1.1$ $[-1.8,-0.4]$) & 31.2 & $-0.5$ $[-1.2,+0.3]$ \\
\addlinespace[3pt]
\multicolumn{6}{@{}l}{\textit{Qwen3-VL-32B$^{\ast}$}} \\
OSWorld-G & $c_{\mathrm{TNZ}}$ (AdaGround) & 76.6 & 76.8 & 79.3 & -- \\
 & fixed $540{\times}540$ crop, $2.78\times$ & 76.4 ($-0.2$) & 77.5 ($+0.7$ $[-1.1,+2.5]$) & 78.0 & $-1.2$ $[-2.9,+0.2]$ \\
\addlinespace[3pt]
ScreenSpot-Pro & $c_{\mathrm{TNZ}}$ (AdaGround) & 72.4 & 75.6 & 75.9 & -- \\
 & fixed $540{\times}540$ crop, $2.78\times$ & 75.3 ($+2.9$) & 76.6 ($+0.9$ $[-0.2,+2.1]$) & 77.0 & $+1.1$ $[0.0,+2.3]$ \\
\addlinespace[3pt]
ScreenSpot-v2 & $c_{\mathrm{TNZ}}$ (AdaGround) & 95.4 & 96.0 & 95.4 & -- \\
 & fixed $540{\times}540$ crop, $2.78\times$ & 93.9 ($-1.5$) & 95.1 ($-0.9$ $[-1.6,-0.2]$) & 95.4 & $-0.1$ $[-0.8,+0.6]$ \\
\addlinespace[3pt]
\bottomrule
\end{tabular}}
\caption{\textbf{Fixed-geometry controls}. Each control replaces $c_{\mathrm{TNZ}}$ with another second chain and leaves everything else unchanged. Values are accuracy (\%). In parentheses: the paired difference in points to the same column of the $c_{\mathrm{TNZ}}$ (AdaGround) row, with $95\%$ CI for arbitration. $2.78\times$: linear scale. $^{h}$Held out and pooled over its instructions (Qwen3-VL-8B only). $^{\ast}$FP8 weights.}
\label{tab:controls}
\end{table*}

Table~\ref{tab:controls} tests whether the complementarity that arbitration exploits is specific to TNZ's geometry. The control is a second \emph{fixed} chain at TNZ's median first-stage geometry (Qwen3-VL-8B, three development benchmarks). On OSWorld-G and ScreenSpot-Pro it is not significantly different from TNZ, after arbitration or in the full pipeline. On these two benchmarks it uses $41$ to $47\%$ fewer visual tokens when the single pass the two chains share is counted once. On Qwen3-VL-8B ScreenSpot-Pro the chain alone is above $c_{\mathrm{TNZ}}$ ($+2.7$ $[+0.9,+4.5]$). On ScreenSpot-v2, whose targets are larger, its $540$-pixel window is too narrow (TNZ's median window there is $990$ and $1{,}074$ pixels). It is significantly worse there after arbitration on both backbones, as it is on held-out UI-Vision (Qwen3-VL-8B), whose targets are small. In the full pipeline these losses are no longer significant.

\section{Reproduced Baselines and ScreenSpot-v2}\label{sec:overall}
Table~\ref{tab:main_backbones} gives the overall accuracy on the three development benchmarks, including ScreenSpot-v2, for which MVP reports no results. Its targets are large (median $85$ pixels), and the single pass already reaches $92.1\%$ and $94.7\%$, which leaves refinement little to add. AdaGround ties the fixed $4\times$ crop on both backbones and is not significantly different from the single pass (Table~\ref{tab:paired}).
\begin{table*}[t]
\centering
\footnotesize
\setlength{\tabcolsep}{2.4pt}
\begin{tabular}{@{}l ccc>{\columncolor{table_color!20}}c ccc>{\columncolor{table_color!20}}c cc>{\columncolor{table_color!20}}c@{}}
\toprule
& \multicolumn{4}{c}{\textbf{OSWorld-G}} & \multicolumn{4}{c}{\textbf{ScreenSpot-Pro}} & \multicolumn{3}{c}{\textbf{ScreenSpot-v2}} \\
\cmidrule(lr){2-5}\cmidrule(lr){6-9}\cmidrule(lr){10-12}
\textbf{Backbone} & Single & \makecell{Fixed\\$4\times$} & MVP & AdaGround & Single & \makecell{Fixed\\$4\times$} & MVP & AdaGround & Single & \makecell{Fixed\\$4\times$} & AdaGround \\
\midrule
Qwen3-VL-8B & 64.7 & 70.7 & 69.9/72.7$^{\dagger}$ & \textbf{76.1} & 51.3 & 63.5 & 61.9/65.3$^{\dagger}$ & \textbf{66.2} & \textbf{92.1} & 91.9 & 91.9 \\
Qwen3-VL-32B$^{\ast}$ & 71.8 & 77.1 & 72.0$^{\dagger}$ & \textbf{79.3} & 55.3 & 73.9 & 74.1$^{\ddagger}$/74.0$^{\dagger}$ & \textbf{75.9} & 94.7 & \textbf{95.4} & \textbf{95.4} \\
\bottomrule
\end{tabular}
\caption{\textbf{Overall accuracy (\%) on the full benchmarks}, with one configuration of AdaGround for both backbones and all benchmarks. Best value per backbone and benchmark in \textbf{bold}. MVP~\citep{mvp2026} values are our reproduction with its official code when unmarked, the authors' released per-sample predictions when marked $^{\ddagger}$, and published results when marked $^{\dagger}$. $^{\ast}$FP8 weights.}
\label{tab:main_backbones}
\end{table*}

Table~\ref{tab:app_baselines} gives the reproduced baselines per category.
\begin{table*}[t]
\centering
\footnotesize
\setlength{\tabcolsep}{4pt}
\renewcommand{\arraystretch}{0.95}
\newlength{\baselineslabelw}\settowidth{\baselineslabelw}{\quad + MVP (authors' released predictions)}
\begin{tabular*}{0.86\textwidth}{@{}w{l}{\baselineslabelw}@{\extracolsep{\fill}}cccccc@{}}
\toprule
\textbf{OSWorld-G} & \textbf{Text} & \textbf{Element} & \textbf{Layout} & \textbf{Fine-gr.} & \textbf{Refusal} & \textbf{Overall} \\
\midrule
Qwen3-VL-8B single pass & 78.2 & 68.5 & 71.9 & 53.7 & 0.0 & 64.7 \\
\quad + fixed $4\times$ crop ($c_{4\times}$ alone) & 80.1 & 77.3 & 78.7 & 58.4 & 0.0 & 70.7\gain{6.0} \\
\quad + MVP (official code\smash{${}^{\ddagger}$}, reproduced) & 78.2 & 76.4 & 79.4 & 56.4 & 0.0 & 69.9\gain{4.8} \\
\quad + AdaGround & 85.1 & 80.0 & 81.8 & 65.1 & 27.8 & 76.1\gain{11.3} \\
\midrule
Qwen3-VL-32B\smash{$^{\ast}$} single pass & 81.6 & 75.8 & 77.5 & 63.1 & 0.0 & 71.8 \\
\quad + fixed $4\times$ crop ($c_{4\times}$ alone) & 82.0 & 84.2 & 86.6 & 63.8 & 0.0 & 77.1\gain{5.3} \\
\quad + AdaGround & 85.8 & 84.8 & 84.6 & 67.1 & 14.8 & 79.3\gain{7.4} \\
\bottomrule
\end{tabular*}

\vspace{6pt}
\begin{tabular*}{0.86\textwidth}{@{}w{l}{\baselineslabelw}@{\extracolsep{\fill}}ccccccc@{}}
\toprule
\textbf{ScreenSpot-Pro} & \textbf{Dev.} & \textbf{Creative} & \textbf{CAD} & \textbf{Scientific} & \textbf{Office} & \textbf{OS} & \textbf{Overall} \\
\midrule
Qwen3-VL-8B single pass & 47.2 & 46.0 & 44.4 & 53.9 & 71.3 & 49.0 & 51.3 \\
\quad + fixed $4\times$ crop ($c_{4\times}$ alone) & 60.5 & 54.3 & 60.5 & 62.6 & 83.9 & 65.3 & 63.5\gain{12.2} \\
\quad + MVP (official code, reproduced) & 58.2 & 54.5 & 57.5 & 60.6 & 82.6 & 63.8 & 61.9\gain{10.4} \\
\quad + AdaGround & 61.5 & 57.2 & 65.5 & 66.9 & 85.2 & 66.3 & 66.2\gain{14.9} \\
\midrule
Qwen3-VL-32B\smash{$^{\ast}$} single pass & 43.1 & 53.7 & 57.1 & 63.0 & 73.0 & 43.9 & 55.3 \\
\quad + fixed $4\times$ crop ($c_{4\times}$ alone) & 69.9 & 67.7 & 74.3 & 71.3 & 90.0 & 74.5 & 73.9\gain{18.5} \\
\quad + MVP (authors' released predictions) & 71.6 & 69.5 & 74.7 & 70.5 & 87.8 & 74.0 & 74.1 \\
\quad + AdaGround & 71.6 & 70.1 & 76.6 & 74.8 & 91.3 & 75.0 & 75.9\gain{20.6} \\
\bottomrule
\end{tabular*}

\vspace{6pt}
\begin{tabular*}{0.86\textwidth}{@{}w{l}{\baselineslabelw}@{\extracolsep{\fill}}cccc@{}}
\toprule
\textbf{UI-Vision} (held out) & \textbf{Basic} & \textbf{Func.} & \textbf{Spatial} & \textbf{Avg.} \\
\midrule
Qwen3-VL-8B single pass & 29.2 & 30.1 & 11.8 & 23.7 \\
\quad + fixed $4\times$ crop ($c_{4\times}$ alone) & 33.6 & 35.4 & 15.5 & 28.2\gain{4.5} \\
\quad + AdaGround & 37.0 & 38.7 & 20.4 & 32.0\gain{8.3} \\
\midrule
Qwen3-VL-32B\smash{$^{\ast}$} single pass & 44.2 & 43.2 & 22.6 & 36.7 \\
\quad + fixed $4\times$ crop ($c_{4\times}$ alone) & 50.5 & 50.5 & 30.6 & 43.9\gain{7.2} \\
\quad + AdaGround & 52.5 & 51.7 & 38.4 & 47.6\gain{10.9} \\
\bottomrule
\end{tabular*}
\caption{\textbf{Reproduced baselines}, accuracy (\%) per category, which the main tables omit. The fixed $4\times$ crop is AdaGround's chain $c_{4\times}$ alone (an ablation, Section~\ref{sec:components}). MVP~\citep{mvp2026} is reproduced with its official code on Qwen3-VL-8B, or scored from the per-sample predictions released by its authors for Qwen3-VL-32B ScreenSpot-Pro. ${}^{\ddagger}$On OSWorld-G, MVP runs its ScreenSpot-Pro pipeline (Appendix~\ref{sec:efficiency_app}). Single pass and AdaGround as in Tables~\ref{tab:osworld_g}, \ref{tab:main_sspro} and~\ref{tab:uivision}. {\color{gaincolor}Subscripts}: gain over the method's own single pass. MVP's single pass scores $65.1$ and $51.5$ in our reproduction, and the released predictions include none. UI-Vision Avg.: mean of the three tasks. $^{\ast}$FP8 weights.}
\label{tab:app_baselines}
\end{table*}

\section{MVP Comparison and Statistical Notes}
Table~\ref{tab:adaground_vs_mvp} contrasts the designs of AdaGround and MVP~\citep{mvp2026}.
\begin{table}[t]
\centering
\footnotesize
\setlength{\tabcolsep}{3pt}
\begin{tabular}{@{}>{\raggedright\arraybackslash}p{0.22\linewidth}>{\raggedright\arraybackslash}p{0.35\linewidth}>{\raggedright\arraybackslash}p{0.35\linewidth}@{}}
\toprule
 & \textbf{MVP}~\citep{mvp2026} & \textbf{AdaGround (ours)} \\
\midrule
Target scale & not modelled; fixed-size attention crops & estimated from the model's own box; $k$ and $d$ solved per sample (TNZ) \\
View selection & attention maps from a chosen layer & size estimate from the model's own box output \\
Aggregation & spatial clustering of 3 predictions, largest group & verifier arbitration over a static chain portfolio; verification-gated acceptance \\
Failure handling & none & negative-feedback re-grounding; existence-gated abstention; compositional cell grounding \\
Model access & attention maps (white box) & black box: text and first-token probabilities \\
Decoding & greedy & greedy \\
\bottomrule
\end{tabular}
\caption{\textbf{AdaGround vs.\ MVP}: design differences.}
\label{tab:adaground_vs_mvp}
\end{table}

\subsection{MVP Setup}\label{sec:efficiency_app}
On Qwen3-VL-8B we reproduce MVP with its official code, prompt and published configuration (attention layer $24$, two sub-image inferences) under Hugging Face transformers~\citep{wolf2020transformers}. The only change, made for speed, turns the vision tower's patch-embedding convolution into the equivalent linear map. On OSWorld-G we use its ScreenSpot-Pro attention pipeline, because its released OSWorld-G code implements a different procedure. The reproduction scores below the published results (Table~\ref{tab:main_backbones}). So does MVP's own single pass in our environment, with $65.1\%$ and $51.5\%$ on OSWorld-G and ScreenSpot-Pro against the published $68.8\%$ and $55.0\%$. The gain MVP adds to its single pass is reproduced, at $+4.8$ and $+10.4$ points against the published $+3.9$ and $+10.3$. On Qwen3-VL-32B we score the ScreenSpot-Pro predictions released by MVP's authors, which allow a paired test, and list the published OSWorld-G result.

\subsection{Scoring and Paired Tests}\label{sec:statistics}
\paragraph{Scoring.} OSWorld-G uses its released evaluation, which scores polygon targets point-in-polygon and credits an absent target only on abstention. The other benchmarks are scored point-in-box with abstention off. As in the official scorer, an answer without a parsable point is wrong. On $12$ of the $54$ absent-target instructions, Qwen3-VL-32B states in text, in every configuration, that the element is not on the screen, and it returns no coordinate within the $32$-token limit of a grounding pass. Only Eq.~\eqref{eq:abstain} counts as a refusal, so these $12$ are wrong for every method. The rows of Table~\ref{tab:osworld_g} marked~${}^{\circ}$ are quoted from \citet{osworldg2024}, which evaluated them on the original instructions. On the refined instructions, its Figure~5 reports $63.7$ for UI-TARS-72B, $57.5$ for Operator, $47$ for Gemini-2.5-pro and $63.8$ for Jedi-7B.

\paragraph{Paired tests.} Paired differences, in Table~\ref{tab:paired} and Appendix~\ref{sec:app_ablation}, carry $95\%$ intervals from $4{,}000$ bootstrap resamples of screenshots (the samples of one screenshot resampled together). For a baseline with abstention (superscript~$b$), the verifier scores the baseline's own answer, with the same existence query as AdaGround.
\begin{table*}[t]
\centering
\footnotesize
\setlength{\tabcolsep}{3.6pt}
\renewcommand{\arraystretch}{0.95}
\begin{tabular}{@{}lccccccc@{}}
\toprule
\textbf{Benchmark} & vs.\ single & vs.\ fixed $4\times$ & vs.\ fixed $4\times^{a}$ & vs.\ fixed $4\times^{b}$ & vs.\ MVP & vs.\ MVP$^{a}$ & vs.\ MVP$^{b}$ \\
\midrule
\multicolumn{8}{@{}l}{\textit{Qwen3-VL-8B}} \\
OSWorld-G & \makecell{$+11.3$\\$[+8.3,+14.4]$} & \makecell{$+5.3$\\$[+2.3,+8.5]$} & \makecell{$+2.8$\\$[+0.4,+5.4]$} & \makecell{$+2.3$\\$[0.0,+4.8]$} & \makecell{$+6.2$\\$[+3.3,+9.1]$} & \makecell{$+3.7$\\$[+1.3,+6.1]$} & \makecell{$+3.0$\\$[+0.6,+5.4]$} \\
 & 71/7 & 47/17 & 32/16 & 32/19 & 53/18 & 38/17 & 38/21 \\[2pt]
ScreenSpot-Pro & \makecell{$+14.9$\\$[+12.8,+16.9]$} & \makecell{$+2.7$\\$[+1.3,+4.0]$} & -- & -- & \makecell{$+4.2$\\$[+2.6,+5.9]$} & -- & -- \\
 & 276/41 & 81/39 &  &  & 125/58 &  &  \\[2pt]
ScreenSpot-v2 & \makecell{$-0.2$\\$[-1.5,+1.0]$} & \makecell{$0.0$\\$[-1.0,+1.0]$} & -- & -- & -- & -- & -- \\
 & 31/34 & 22/22 &  &  &  &  &  \\[2pt]
UI-Vision & \makecell{$+8.3$\\$[+7.2,+9.4]$} & \makecell{$+3.9$\\$[+2.9,+4.8]$} & -- & -- & -- & -- & -- \\
 & 606/150 & 388/175 &  &  &  &  &  \\[2pt]
\midrule
\multicolumn{8}{@{}l}{\textit{Qwen3-VL-32B$^{\ast}$}} \\
OSWorld-G & \makecell{$+7.4$\\$[+4.8,+10.1]$} & \makecell{$+2.1$\\$[-0.2,+4.4]$} & \makecell{$+0.7$\\$[-1.3,+2.7]$} & \makecell{$+0.7$\\$[-1.4,+2.7]$} & -- & -- & -- \\
 & 51/9 & 28/16 & 20/16 & 21/17 &  &  &  \\[2pt]
ScreenSpot-Pro & \makecell{$+20.6$\\$[+18.3,+22.8]$} & \makecell{$+2.0$\\$[+0.7,+3.4]$} & -- & -- & \makecell{$+1.8$\\$[+0.3,+3.3]$} & -- & -- \\
 & 358/33 & 76/44 &  &  & 92/64 &  &  \\[2pt]
ScreenSpot-v2 & \makecell{$+0.7$\\$[-0.2,+1.7]$} & \makecell{$0.0$\\$[-0.9,+0.9]$} & -- & -- & -- & -- & -- \\
 & 25/16 & 16/16 &  &  &  &  &  \\[2pt]
UI-Vision & \makecell{$+11.0$\\$[+9.6,+12.5]$} & \makecell{$+3.8$\\$[+2.7,+5.0]$} & -- & -- & -- & -- & -- \\
 & 820/216 & 508/299 &  &  &  &  &  \\[2pt]
\bottomrule
\end{tabular}
\caption{\textbf{Paired differences} (points) of AdaGround on the full benchmarks, with $95\%$ confidence intervals (Appendix~\ref{sec:statistics}) and, below, samples won/lost. MVP: our reproduction (Qwen3-VL-8B) or the authors' released predictions (Qwen3-VL-32B ScreenSpot-Pro). UI-Vision: held out, pooled over its instructions. $^{a}$AdaGround without abstention. $^{b}$The baseline with the abstention rule of Eq.~\eqref{eq:abstain} applied to its own answer (OSWorld-G). $^{\ast}$FP8 weights.}
\label{tab:paired}
\end{table*}

\section{Token-Span Analysis}\label{sec:app_tokenspan}
Unless stated otherwise, this appendix uses the following baseline passes of Qwen3-VL-8B: the single pass, and a square crop of side $d_0(W,H)$ centred on its answer, rendered at native resolution (the \emph{square crop}) and at $4\times$ area (the \emph{square $4\times$ crop}). These passes are separate from the experiments of Section~\ref{sec:exp} and have their own crop geometry and preprocessing, so their accuracies differ from Table~\ref{tab:main_backbones}.

\subsection{Definition, Relation to Area Measures, and Pitch Sensitivity}
The token span $\sigma = sk/\pi$ (Eq.~\eqref{eq:span}, with $\pi=32$ for Qwen3-VL~\citep{qwen3vl2025}) is measured in the rendered view. An area measure, the \emph{effective tokens on the target} $r_{\mathrm{eff}} = T(A_I)\cdot A_t/A_I$, is defined on the whole screenshot instead. Here $A_t$ and $A_I$ are the areas of the target and the screenshot, and $T(A_I)$ is the number of visual tokens of the whole screenshot. It satisfies $r_{\mathrm{eff}}\approx\sigma^2$ for a square target at $k=1$. A $4\times$ up-sampled crop leaves $A_t/A_I$ unchanged but doubles $\sigma$. This is why $\sigma$, and not the area ratio, predicts the effect of a zoom. For example, a $24$-pixel target on a $3840{\times}2160$ screenshot spans $0.4$ tokens on TNZ's coarse view ($k\approx0.5$), $0.75$ at $k=1$ and $1.5$ after a $4\times$ crop. Eq.~\eqref{eq:k} with the constants of Table~\ref{tab:constants} solves $k=\operatorname{clip}(3{\cdot}32/24,1,4)=4$ and lands the target on $3.0$.

\paragraph{Pitch sensitivity.} The pitch $\pi$ only fixes the unit. What matters is that size is measured \emph{after} rendering. A gate on original pixels is a different rule, because the same target is rendered at different scales on different screens.

\subsection{Accuracy versus Token Span}\label{sec:app_law}
\begin{figure}[t]
\centering
\includegraphics[width=\columnwidth]{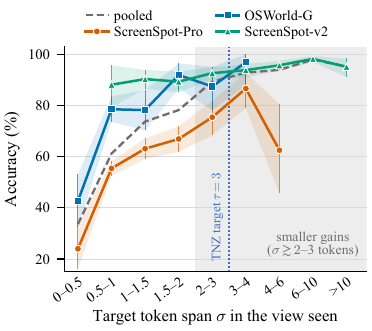}
\caption{\textbf{Accuracy versus token span} per benchmark (bins with $\ge30$ views) and pooled (dashed), on views that contain the target (Table~\ref{tab:target_size}). Whiskers and bands: $95\%$ screenshot-bootstrap intervals. Dotted line: TNZ's target span $\tau=3$.}
\label{fig:size_and_mech}
\end{figure}

\begin{figure*}[t]
\centering
\includegraphics[width=0.8\textwidth]{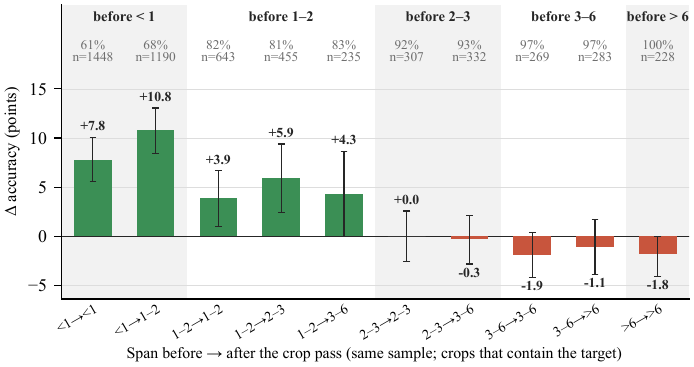}
\caption{\textbf{Within-sample change from the crop pass} (Table~\ref{tab:target_size}, bottom), by span before~$\to$~after (green gain, red loss). Above each bar: accuracy before the crop and $n$. Whiskers: $95\%$ paired screenshot-bootstrap intervals.}
\label{fig:within_sample}
\end{figure*}

Table~\ref{tab:target_size} and Figure~\ref{fig:size_and_mech} pool the whole-screen and crop passes of the three baselines on the three benchmarks (nine passes). They are restricted to the $15{,}488$ of $16{,}813$ view--sample pairs whose view contains the target, so that a miss measures legibility and not framing. Accuracy climbs steeply below about three tokens and gains much less above. In the pooled data, all four steps below three tokens exclude zero, although the step from $1$--$1.5$ to $1.5$--$2$ tokens does so only narrowly ($+4.5$ $[+0.03,+8.7]$). Two paired contrasts isolate the span. Up-sampling the same crop $4\times$, which changes $\sigma$ alone, gains $2.6$ $[+1.1,+4.0]$, $6.6$ $[+4.4,+8.7]$ and $0.9$ $[-0.1,+2.0]$ points on ScreenSpot-Pro, OSWorld-G and ScreenSpot-v2. The second contrast compares the same sample on the coarse view and on the crop of the same baseline (bottom of Table~\ref{tab:target_size} and Figure~\ref{fig:within_sample}). The crop gains $+7.8$ and $+10.8$ points below one token and nothing significant once the target spans two tokens or more. We call this empirical trend on three benchmarks with Qwen3-VL-8B the \emph{token-span regularity}. It governs the \emph{benefit} of a zoom.

\begin{table*}[t]
\centering
\footnotesize
\setlength{\tabcolsep}{3.6pt}
\begin{tabular}{l rr rr rr rr}
\toprule
& \multicolumn{2}{c}{\textbf{ScreenSpot-Pro}} & \multicolumn{2}{c}{\textbf{OSWorld-G}} & \multicolumn{2}{c}{\textbf{ScreenSpot-v2}} & \multicolumn{2}{c}{\textbf{Pooled}} \\
\cmidrule(lr){2-3}\cmidrule(lr){4-5}\cmidrule(lr){6-7}\cmidrule(lr){8-9}
$\sigma$ (tokens) & $n$ & Acc. & $n$ & Acc. & $n$ & Acc. & $n$ & Acc.\ [95\% CI] \\
\midrule
$0$--$0.5$ & 342 & 24.0 & 366 & 42.6 & 0 & -- & 708 & 33.6 [26.7, 40.4] \\
$0.5$--$1$ & 4180 & 55.4 & 1109 & 78.5 & 185 & 88.1 & 5474 & 61.2 [58.9, 63.6] \\
$1$--$1.5$ & 1396 & 63.1 & 532 & 78.2 & 743 & 90.3 & 2671 & 73.7 [70.6, 76.6] \\
$1.5$--$2$ & 813 & 66.9 & 173 & 91.9 & 615 & 89.3 & 1601 & 78.2 [75.0, 81.3] \\
$2$--$3$ & 317 & 75.4 & 174 & 87.4 & 1422 & 92.6 & 1913 & 89.3 [86.8, 91.6] \\
$3$--$4$ & 149 & 86.6 & 32 & 96.9 & 817 & 93.8 & 998 & 92.8 [90.3, 95.1] \\
$4$--$6$ & 56 & 62.5 & 27 & 92.6 & 1003 & 95.7 & 1086 & 93.9 [91.6, 96.0] \\
$6$--$10$ & 19 & 89.5 & 11 & 90.9 & 673 & 98.2 & 703 & 97.9 [96.5, 99.0] \\
$>10$ & 2 & 100.0 & 8 & 100.0 & 324 & 95.1 & 334 & 95.2 [91.5, 98.4] \\
\bottomrule
\end{tabular}

\vspace{0.6em}
\begin{tabular}{llrcrrr}
\toprule
\multicolumn{7}{c}{\textbf{Within-sample} ($5{,}390$ pairs in the ten cells with $n\ge40$; $5{,}399$ in all)} \\
\cmidrule(lr){1-7}
Before & After & $n$ & Mean $\sigma$ before$\to$after & Acc.\ before & Acc.\ after & $\Delta$ [95\% CI] \\
\midrule
$<1$ & $<1$ & 1448 & $0.65\to0.70$ & 61.0 & 68.8 & \cellcolor{table_color!20}$+7.8$ $[+5.5, +10.1]$ \\
$<1$ & 1--2 & 1190 & $0.73\to1.45$ & 68.1 & 78.8 & \cellcolor{table_color!20}$+10.8$ $[+8.4, +13.1]$ \\
1--2 & 1--2 & 643 & $1.37\to1.40$ & 82.3 & 86.2 & $+3.9$ $[+1.0, +6.7]$ \\
1--2 & 2--3 & 455 & $1.25\to2.42$ & 81.3 & 87.3 & $+5.9$ $[+2.4, +9.4]$ \\
1--2 & 3--6 & 235 & $1.71\to3.46$ & 83.0 & 87.2 & $+4.3$ $[+0.0, +8.7]$ \\
2--3 & 2--3 & 307 & $2.37\to2.42$ & 92.5 & 92.5 & $+0.0$ $[-2.6, +2.6]$ \\
2--3 & 3--6 & 332 & $2.41\to4.73$ & 93.1 & 92.8 & $-0.3$ $[-2.8, +2.1]$ \\
3--6 & 3--6 & 269 & $3.97\to4.04$ & 97.4 & 95.5 & $-1.9$ $[-4.2, +0.4]$ \\
3--6 & $>6$ & 283 & $4.07\to8.02$ & 96.8 & 95.8 & $-1.1$ $[-3.9, +1.7]$ \\
$>6$ & $>6$ & 228 & $8.40\to12.68$ & 100.0 & 98.2 & \cellcolor{table_color!20}$-1.8$ $[-4.1, +0.0]$ \\
\bottomrule
\end{tabular}
\caption{\textbf{Accuracy versus token span} (Qwen3-VL-8B, with the nine baseline passes of Appendix~\ref{sec:app_tokenspan}), on views that contain the target. Top: accuracy by the token span $\sigma$ in the view the model saw, per benchmark and pooled ($n$ = views). Pooled cells carry $95\%$ screenshot-bootstrap intervals, with all views of a screenshot resampled together. Bottom: the same sample on its coarse view and on the crop of the same baseline, with mean span, accuracy before and after, and the paired change.}
\label{tab:target_size}
\end{table*}

\paragraph{Exceptions and composition.} One ScreenSpot-Pro bin, four to six tokens, runs against the trend ($-24.1$ $[-43.3,-4.3]$ points below the bin beneath it), but it holds only $56$ views from $41$ screenshots. The pooled within-sample fade is partly composition. Low-span cells are $72\%$ ScreenSpot-Pro, and high-span cells are almost all ScreenSpot-v2. Even so, each benchmark follows the trend over its own range (cells with $n\ge40$). Every ScreenSpot-Pro cell starts below two tokens and gains ($+10.6$ to $+13.3$ points). No OSWorld-G cell excludes zero ($0.0$ to $+5.1$). Eight of ten ScreenSpot-v2 cells are negative ($-2.8$ to $+5.7$), none significantly. The sign flips from ScreenSpot-Pro to ScreenSpot-v2. The square $4\times$ crop beats the single pass on the former ($+10.7$ $[+8.8,+12.6]$) and loses to it on the latter ($-2.6$ $[-4.0,-1.2]$).

\subsection{The \texorpdfstring{$2{\times}2$}{2x2} Control and ScreenSpot-v2}\label{sec:ssv2_results}
A crop also discards the screen around the target, a \emph{cost} the regularity does not describe. A $2{\times}2$ control over $\{$whole screen, crop$\}\times\{$native, $4\times\}$ places both the benefit and the cost in the crop. Rendering the whole screen at four times the area raises the median view scale from $1.01$ to $1.76$, so the median span grows $1.7\times$ with no context lost. Yet accuracy changes by only $+0.4$ points on ScreenSpot-Pro ($p=0.72$, at $2.2$ times the visual tokens) and $-0.7$ on ScreenSpot-v2 ($p=0.23$). Cropping at native scale changes them by $+8.1$ and $-3.5$ points. Span bought without restricting the view brings no gain. The crop both raises the span and removes candidate locations, and $\sigma$ says which effect dominates. On ScreenSpot-v2, TNZ's loose first-stage criterion fires on $92.5\%$ and $91.8\%$ of samples (Qwen3-VL-8B and 32B). Gated acceptance keeps the zoomed answer on only $60.8\%$ and $58.2\%$, so $c_{\mathrm{TNZ}}$ stays level with the single pass (Table~\ref{tab:ladder}).

\subsection{What a Well-Placed Crop Leaves Behind}\label{sec:app_residual}
On ScreenSpot-Pro, an oracle square crop centred on the ground truth resolves only $287$ of the single pass's $787$ failures and breaks $50$ correct answers. With the square $4\times$ crop, $322$ of the $618$ remaining failures ($52\%$) have the target inside the crop. In these cases the model saw the target and committed to another location. Refinement therefore needs a signal to accept or reject a zoomed answer, push the model off a wrong choice and choose between disagreeing views.

\subsection{Targets without Visual Evidence}\label{sec:app_noevidence}
Two classes of OSWorld-G instruction are not about legibility. A spreadsheet cell is named by its column and row, whose headers lie outside any crop tight enough to magnify it. On these instructions the zoomed chain solves fewer than the single pass (Appendix~\ref{sec:extended_ablation}). The $54$ instructions whose target is absent are credited only on abstention, so a pipeline that always answers forfeits $9.6$ points however well it grounds. Both call for a decision outside the zoom.

\subsection{From the Analysis to the Design}\label{sec:app_design_rule}
(1)~The steep climb below about three tokens sets the span $\tau$ of every zoom (Eq.~\eqref{eq:k}). (2)~A crop gains most below one token and nothing significant above three, so a missed zoom costs more than an unnecessary one. The first stage therefore fires under the looser criterion $\hat\sigma<\lambda\tau$, $\lambda>1$, and gated acceptance undoes unnecessary zooms. (3)~The span is unobservable, so TNZ estimates it from the model's box. The context a crop discards is not predicted by the span (Appendix~\ref{sec:ssv2_results}), so every zoom is verified. (4)~Errors with the target in view (Appendix~\ref{sec:app_residual}) motivate negative-feedback re-grounding, and (5)~targets without visual evidence (Appendix~\ref{sec:app_noevidence}) motivate CSG and abstention.

\end{document}